\documentclass{article}

 \usepackage[preprint]{neurips_2026}

\usepackage[utf8]{inputenc} 
\usepackage[T1]{fontenc}    
\usepackage{hyperref}       
\usepackage{url}            
\usepackage{booktabs}       
\usepackage{amsfonts}       
\usepackage{nicefrac}       
\usepackage{microtype}      
\usepackage{xcolor}         
\usepackage{float}
\usepackage{amsmath}
\usepackage{bbm}
\usepackage{graphicx}
\usepackage{subcaption}
\usepackage{booktabs}
\usepackage[table]{xcolor}
\usepackage{multirow}
\usepackage{bm}
\usepackage{wrapfig}
\usepackage{tabularx}
\usepackage{caption}
\usepackage{rotating}

\title{Evaluation Metrics for Safe Reinforcement Learning}

\author{%
  Lindsay Spoor\,\thanks{corresponding author: \texttt{l.j.spoor@liacs.leidenuniv.nl}}\,\,\,\thanks{Leiden Institute of Advanced Computer Science, Leiden University, Leiden, the Netherlands} \\
  \And
  Aske Plaat\footnotemark[2] \\
  \And
  Thomas Moerland\footnotemark[2] \\
}

\begin{document}

\maketitle
\setcounter{footnote}{0}

\begin{abstract}

    Safe reinforcement learning (RL) is commonly formalized as a Constrained Markov Decision Process (CMDP), in which an agent maximizes expected reward while keeping its expected cumulative cost below a specified safety bound. Existing safe RL benchmarks predominantly report whether an algorithm is safe on average, following this expectation-based guarantee. We argue that this convention is insufficient to reliably characterize an algorithm's true safety: it fails to capture how often and how severely the safety bound is violated, whether this holds consistently across tasks and safety bounds, and whether training-time behavior is representative of behavior of the final converged policy. Therefore, we introduce (i) \textit{evaluation metrics} for safe RL that address each of these concerns and in addition allow for aggregation across tasks and safety bounds. We furthermore define (ii) a \textit{safety tier system} to systematically categorize and compare algorithms in terms of safety and reliability at both training and for a final policy. Using this framework, we provide (iii) an \textit{empirical safety evaluation} across multiple safety navigation tasks. Our results show that aggregate metrics, distributional reporting, and task- and safety bound-specific results each reveal information the other metrics cannot. We therefore recommend reporting all three jointly, rather than compressing this information into a single value, as is common practice. We provide \href{https://github.com/lindsayspoor/SafeRLEval}{\textit{SafeRLEval\footnote{\url{https://github.com/lindsayspoor/SafeRLEval}}}}, an open-source evaluation suite to support the reliable characterization of safety in future safe RL research.
\end{abstract}

\section{Introduction}\label{sec: Introduction}

Reinforcement learning (RL) addresses sequential decision-making problems by enabling agents to learn from feedback in the form of rewards, with the goal of maximizing their long-term cumulative reward \citep{sutton_reinforcement_nodate}. Despite their success on tasks without critical safety concerns \citep{mnih2015human, schulman_proximal_2017, haarnoja2018soft}, agents deployed in safety-critical domains, such as autonomous driving, robotics, and power systems \citep{ramanujam_safeor-gym_2025}, must often satisfy restrictive constraints on their behavior, in addition to maximizing task performance.

Safe RL provides a framework for such safety-critical tasks, in which the learning objective is extended to explicitly incorporate constraints. This is commonly formalized as a Constrained Markov Decision Process (CMDP) \citep{altman_constrained_1999}, in which the agent's objective is to maximize expected reward subject to $\mathbb{E}[C] \leq d$, where $C$ denotes the cumulative episodic cost and $d$ the safety bound. Safe RL has been extensively studied over the last decade, leading to a variety of approaches \citep{garcia_comprehensive_2015}, including Lagrangian (primal-dual) methods \citep{ray_benchmarking_nodate, tessler_reward_2018, stooke_responsive_2020}, trust-region methods \citep{zhang_first_2020, achiam_constrained_2017}, and increasing cost penalty coefficients in the objective \citep{zhang2022penalized}.

Existing safe RL benchmarks generally report whether an algorithm is safe \textit{on average} \citep{ray_benchmarking_nodate,ji_omnisafe_nodate}. This convention follows directly from the CMDP formulation itself, with the standard formal guarantee, $\mathbb{E}[C]\leq d$, being expectation-based. \citet{ray_benchmarking_nodate} proposed an evaluation protocol that distinguishes average performance during training from that of the final policy, alongside an average training regret capturing cumulative unsafe behavior incurred while learning. However, this protocol does not indicate how often, or how severely, a policy violates its constraints, nor does it distinguish between evaluating the final policy with exploration noise enabled and evaluating it as a greedy policy, i.e., with exploration noise disabled. Even though more recent benchmark suites \citep{ji_safety-gymnasium_2024, ji_omnisafe_nodate, tomilin2026crax, ramanujam_safeor-gym_2025} have substantially expanded the range of available tasks, algorithms, and constraint types, the standard evaluation protocol has changed comparatively little. We argue that an expectation-based guarantee, with a measure of safety collapsed into a single aggregate value, remains insufficient to reliably characterize a method's true safety. 

\begin{figure}[t]
    \centering
    \includegraphics[width=0.7\textwidth]{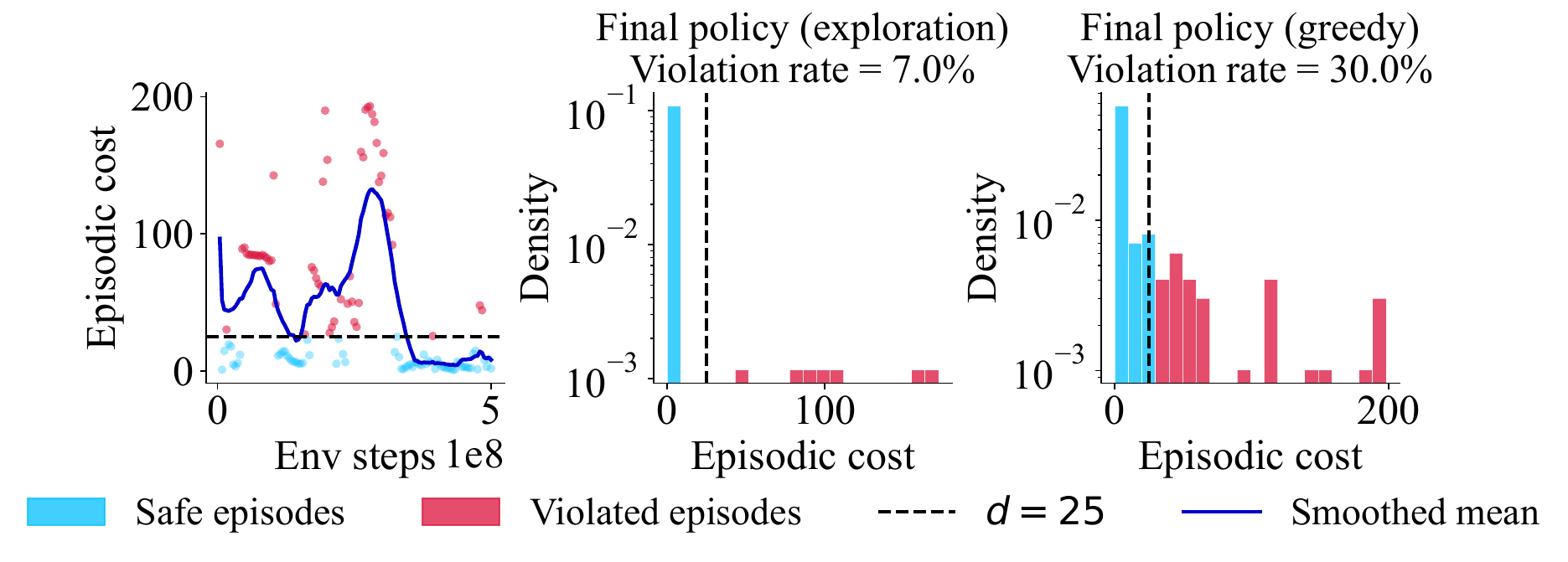}
    \caption{An example illustrating that the formal guarantee $\mathbb{E}[C]\leq d$ in the CMDP framework falls short of reliably ensuring safety, from a single-seed run of Lagrangian-based PPO, task: Safe Circle Point (Level 1), $d=25$ \citep{tomilin2026crax}. (left) The Monte Carlo estimate of $\mathbb{E}[C]$ is shown by the smoothed mean. All individual data points show single-episode costs. Even though the final policy with exploration noise shows that the full distribution of episodes lies below $d$ (middle), greedy policy rollouts of the final policy show a distribution with a substantially higher violation rate (right).}
    \label{fig:motivation}
\end{figure}

Additionally, behavior observed during training is often subject to, and optimized under, exploration noise. However, the performance can differ for the behavior of a final, converged policy evaluated without exploration noise, since such a policy is commonly deployed deterministically in simulation. Figure~\ref{fig:motivation} illustrates this on a representative single-seed training run of Lagrangian PPO \citep{tomilin2026crax}: although $93\%$ of evaluated episodes of the converged final policy lie below the bound $d$ when exploration noise is enabled, a substantial fraction of episodes violate $d$ when the same policy is instead evaluated deterministically, showing a distributional shift between the two settings that, in this case, results in an even higher violation rate. Such shifts can occur because Lagrangian methods only constrain the expected cost under the stochastic policy used during training, while the deterministic policy, which simply takes the most likely action at each step, was never directly subject to this constraint.

This motivates the need to unpack the aggregate value of safety, typically reported as cost on average, into more informative measures, such as violation rate, violation magnitude, and cost deviation from the safety bound, both with and without exploration noise in the final policy. Moreover, current evaluation practices in safe RL have yet to adopt statistically rigorous aggregation of performance across a range of tasks, as well as across a range of safety bounds \citep{spoor2025towards}. \citet{agarwal2021deep} showed in the general RL literature that reporting single point estimates of performance, without accounting for variability across runs and tasks, can produce misleading comparisons, and proposed more statistically robust aggregate measures. Such aggregation across tasks, and in particular across safety bounds, remains largely absent from the safe RL literature. Moreover, Safe RL yet has to adopt an evaluation protocol that jointly reports these distributional properties alongside the single-value metrics of evaluated algorithms.

We therefore propose the following contributions: 
\begin{enumerate}
    \item \textbf{Evaluation metrics for safe RL}\hspace{1em} We introduce evaluation metrics for safe RL that go beyond reporting average reward and cost alone, focusing on violation frequency, violation magnitude, and cost deviation, each normalized to allow for aggregation across both tasks and safety bounds. We provide an open-source evaluation suite at \url{https://github.com/lindsayspoor/SafeRLEval}.
    \item \textbf{Safety tier system}\hspace{1em} We define a set of safety tiers, based on conditions tied to the proposed evaluation metrics, that allow existing safe RL algorithms to be systematically categorized and compared by their safety and reliability, both for any policy during training and for the final converged policy, rather than by average performance alone.
    \item \textbf{Empirical safety evaluation}\hspace{1em} We provide a robust empirical evaluation across a variety of tasks, multiple safety bounds, a variety of popular safe RL algorithms, and 30 seeds. Alongside the standard practice of reporting reward and cost training curves, our evaluation additionally reports the episodic cost distribution over the entire course of training, for the final policy under exploration noise, and for the final greedy policy, both for individual tasks and safety bounds and aggregated across all such conditions.
\end{enumerate}

\section{Safe reinforcement learning}\label{sec: Safe Reinforcement Learning}

Safe RL commonly uses the formal framework of a Constrained Markov Decision Process (CMDP) \citep{altman_constrained_1999}, an extension of the Markov Decision Process (MDP) \citep{sutton_reinforcement_nodate, bellman_markovian_1957}, in which we not only want to maximize desired behavior, but also want to minimize undesired, unsafe behavior. A CMDP is defined as the tuple $\mathcal{M}=(\mathcal{S,A}, P, R, P_0, \gamma, \mathcal{C})$, where $\mathcal{S}$ and $\mathcal{A}$ are the set of all possible states and actions, respectively, $P$ is the transition dynamics distribution $P:\mathcal{S}\times\mathcal{A}\rightarrow \Delta(\mathcal{S})$, $R$ is the reward function $R: \mathcal{S \times A \times S}\rightarrow \mathbb{R}$, $P_0\in\Delta(\mathcal{S})$ is the initial state distribution, and $\gamma \in [0,1)$ is the discount factor. A set of $m$ cost functions $\mathcal{C}:=\{C_1, ..., C_m\}$ with associated safety bounds $\mathcal{D}:=\{d_1,...,d_m\}$ maps transition tuples to costs, $\mathcal{C:S\times A\times S}\rightarrow \mathbb{R}^m$. Actions are selected from a policy $\pi: \mathcal{S}\rightarrow\Delta(\mathcal{A})$, $s\mapsto \pi(\cdot|s)$. For a trajectory $\tau=(s_t, a_t, s_{t+1}, ...)$, the discounted return and discounted cost are given by $R(\tau) = \sum_{t=0}^{T}\gamma^t R(s_{t}, a_{t}, s_{t+1})$ and $C_i(\tau) = \sum_{t=0}^{T}\gamma^t C_i(s_{t}, a_{t}, s_{t+1})$, respectively.

The objective of a CMDP is to find a policy that maximizes expected discounted return while keeping each expected discounted cost within its safety bound:
\begin{eqnarray} \label{eq: optimization problem CMDP}
\begin{aligned}
    &\max_{\pi \in \Pi} \; \mathbb{E}_{\tau\sim\pi,P}\big[R(\tau)\big] \\
    &\text{s.t.} \quad \mathbb{E}_{\tau\sim\pi,P}\big[C_i(\tau)\big] \leq d_i, \quad i=1,...,m,
\end{aligned}
\end{eqnarray}
where $\Pi$ denotes the set of policies. Unlike standard MDPs \citep{bellman_markovian_1957}, policy search under a CMDP must additionally ensure that each policy update remains feasible with respect to its constraints, i.e., that it stays within the feasible set $\Pi_{\mathcal{C}} := \{\pi\in\Pi: \mathbb{E}_{\tau\sim\pi}[C_i(\tau)] \leq d_i \; \forall i\}$, rather than optimizing over $\Pi$ alone. For a single cost function ($m=1$), we write this constraint as $\mathbb{E}[C] \leq d$, with $C$ the cumulative episodic cost and $d$ the safety bound. 

In this study, we evaluate several widely used safe RL algorithms, introduced below. A large family of safe RL algorithms address the constrained optimization problem in Eq.~\ref{eq: optimization problem CMDP} by modifying the underlying objective function itself. Lagrangian relaxation \citep{ray_benchmarking_nodate, tessler_reward_2018, stooke_responsive_2020} penalizes the reward objective in proportion to the constraint violation, with the Lagrange multiplier learned jointly with the policy via dual gradient ascent. Under convexity conditions, this guarantees feasibility only asymptotically, offering no guarantee for any individual policy iterate. P3O \citep{zhang2022penalized} similarly penalizes the objective, but instead progressively increases a fixed-form penalty coefficient in response to observed violations, avoiding the instability of a learned Lagrange multiplier at the loss of a formal convergence guarantee. An alternative family of methods instead restricts the policy update itself. FOCOPS \citep{zhang_first_2020} solves this constrained update via a first-order approximation. This approach, in approximation, enforces feasibility at every policy iteration.


\section{Evaluation metrics}\label{sec: Evaluation Metrics}

We first characterize an algorithm's safety within a single fixed condition: a single task $m$, safety bound $d$, algorithm $a$, and seed $k$. Each of the introduced metrics can be used to evaluate a fixed policy $\pi_j$ at policy iterate $j\in\{0,...,J\}$, where $J$ denotes the final policy iterate, resulting in the converged policy $\pi_{\text{final}}$. Instance $i$ denotes a single evaluation episode out of $N$ episodes evaluated under policy $\pi_j$, with $N$ episodes forming a Monte Carlo estimate of the expectation in the CMDP definition from Eq.~\ref{eq: optimization problem CMDP}. We then describe how these metrics are aggregated across all policy iterates, tasks, and safety bounds, and furthermore introduce a safety tier system based on the resulting metrics, enabling a generalized comparison of algorithms.

\paragraph{Exploration noise} Exploration noise is an inherent part of the training process itself, making it the relevant condition for characterizing safety while the policy is still being learned, whereas a converged policy is typically deployed deterministically in practice, making the greedy evaluation the relevant condition for characterizing deployment-time safety. All of our proposed metrics can be evaluated under two distinct policy conditions: with exploration noise enabled, or with a greedy policy, with exploration noise disabled. Comparing them directly isolates the effect that exploration noise itself has on the safety of an algorithm. Throughout this paper, we report metrics evaluated with exploration noise enabled when characterizing behavior over the course of training, and metrics evaluated under both a greedy and an exploration-based policy when characterizing the final policy, for direct comparison.

\subsection{Metrics for a fixed policy}

As a standard performance metric already reported by prior benchmarks, the \textbf{average reward and cost} are described as in Eq. \ref{eq: baseline perf}.
\begin{align} \label{eq: baseline perf}
  \bar R^j_{(m,d,a,k)} &= \frac{1}{N}\sum_{i=1}^{N} r_{i,(m,d,a,k)}, \qquad
  \bar C^j_{(m,d,a,k)} = \frac{1}{N}\sum_{i=1}^{N} c_{i,(m,d,a,k)},
\end{align}
where $r_{i,(m,d,a,k)}$ and $c_{i,(m,d,a,k)}$ are the reward and cost of rollout episode $i$ under policy iterate $j$, for seed $k \in \{1,\dots,K\}$, task $m$, safety bound $d$, and algorithm $a$.

We propose the \textbf{violation rate} as a metric that indicates how often an algorithm violates constraints, which is the fraction of episodes on task $m$ with bound $d$ in which algorithm $a$ exceeded the safety bound $d$, as described by Eq. \ref{eq: viol rate}.
\begin{align} \label{eq: viol rate}
  V^j_{(m,d,a,k)} = \frac{1}{N}\sum_{i=1}^{N} \mathbbm{1}[c_{i,(m,d,a,k)} > d],
\end{align}
where $\mathbbm{1}[\cdot]$ is the indicator function.

Additionally, we introduce the \textbf{mean normalized cost deviation} as the metric that captures how far algorithm $a$'s average cost sits above ($>0$, unsafe) or below ($<0$, safe) the bound, scaled by the bound itself, as in Eq.~\ref{eq: mean norm cost dev}.
\begin{align} \label{eq: mean norm cost dev}
  D^j_{\text{norm},(m,d,a,k)} = \frac{\bar C^j_{(m,d,a,k)} - d}{d}.
\end{align}
Unlike \citet{ray_benchmarking_nodate}, who normalize deviation from the bound relative to a naive, unconstrained baseline algorithm like PPO \citep{schulman_proximal_2017}, we normalize directly by the safety bound $d$ itself. Because an unconstrained baseline typically incurs costs far above $d$, the deviation term would be large, and the normalized scores of safe RL algorithms would vanish.
We instead treat $d$ as the natural unit of measurement for cost deviation, centering $D_{\text{norm}}$ around $0$ for $\bar C = d$.

Among the episodes in which algorithm $a$ violated safety bound $d$, the \textbf{normalized violation magnitude} then captures how much they overshot the bound on average, relative to the bound itself, following Eq.~\ref{eq: dnorm+}.
\begin{align} \label{eq: dnorm+}
  D^{j}_{\text{norm}^+,(m,d,a,k)} =
  \begin{cases}
    \dfrac{1}{N^{+}_{a,k}\, d}\displaystyle\sum_{i=1}^N \big(c_{i,(m,d,a,k)}-d\big)^{+}
      & \text{if } N^{+}_{a,k} > 0, \\[1.5ex]
    0 & \text{if } N^{+}_{a,k} = 0,
  \end{cases}
\end{align}
where $N^{+}_{a,k} = \sum_{i=1}^N \mathbbm{1}[c_{i,(m,d,a,k)} > d]$ is the number of violating rollout episodes for algorithm $a$ under seed $k$, and $(x)^{+} := \max(x,0)$ denotes the positive part of $x$.

Along with $D_{\text{norm}}$ from Eq. \ref{eq: mean norm cost dev}, which reflects how far the algorithm's cost sits from the bound on average, over all episodes,  $D_{\text{norm}^+}$ in Eq. \ref{eq: dnorm+} is restricted to the set of violating episodes only. This distinction matters because two algorithms with an identical average cost can exhibit very different violation profiles. An algorithm may violate the bound frequently but only slightly on each occasion, or violate rarely but catastrophically when it does.

\subsection{Aggregation across policy iterates} \label{sec: Aggregation across policy iterates}
Each metric $X^j_{(m,d,a,k)}\in\{\bar{R}^j_{(m,d,a,k)},\bar{C}^j_{(m,d,a,k)}, V^j_{(m,d,a,k)}, D^j_{\text{norm},(m,d,a,k)}, D^j_{\text{norm}^+,(m,d,a,k)}\}$ is defined for a fixed policy $\pi_j$ at policy iterate $j$. To evaluate an algorithm's performance across the entire course of training, i.e., aggregated over all policy iterates, we take the average across all iterates, as shown in Eq.~\ref{eq: policy iterate agg}. Since this is meant to capture behavior during the training process, we evaluate policy $\pi_j$ with exploration noise enabled for this aggregation.
\begin{align} \label{eq: policy iterate agg}
    X_{(m,d,a,k)} = \frac{1}{J}\sum_{j=0}^JX^j_{(m,a,d,k)}.
\end{align}

\subsection{Aggregation across tasks and safety bounds} \label{sec:Aggregation}

To characterize an algorithm's overall safety performance, each per-seed metric is first averaged over seeds $k\in\{1,...,K\}$, and then aggregated over the full set of task--bound pairs $(m,d) \in \mathcal{P}$. Following \citet{agarwal2021deep}, we report the interquartile mean (IQM) across $\mathcal{P}$, taking the mean of the values that fall between the 25th and 75th percentile. We denote this as $\text{IQM}\big(X^{j}_{a}\big)$ for a fixed policy $\pi_j$, and as $\text{IQM}\big(X_{a}\big)$ for the metric aggregated over the entire course of training, as defined in Eq.~\ref{eq: policy iterate agg}.

\paragraph{Safety tiers} While the level of detail in the introduced metrics is essential for a reliable evaluation, it can also make it difficult to directly compare algorithms. Therefore, we define thresholds on each $X\in\{V_a, D_{\text{norm},a}, D_{\text{norm}^+,a}\}$, separating algorithms into safety tiers from level 0 to 4, with increasingly stringent safety requirements, following Table~\ref{tab:safety_tiers}. We argue that any algorithm should be assessed on these tiers under two settings: over the course of training, using $\text{IQM}(X^{\text{expl}}_a)$, where the superscript $\text{expl}$ denotes evaluation with exploration noise enabled; and at the final, greedy policy, using $\text{IQM}(X^{\text{final,greedy}}_a)$. This means, for example, that an algorithm may achieve tier 1 during training, but tier 2 at the final greedy policy.

\begin{table}[h]
\centering
\small
\caption{Safety tiers based on threshholds for the aggregated metrics $\text{IQM}(D_{\text{norm}})$, $\text{IQM}(V)$, and $\text{IQM}(D_{\text{norm}^+})$, for a single algorithm $a$. \vspace{0.3em}}
\begin{tabular}{clll}
\toprule
\textbf{Tier} & \textbf{Label} & \textbf{Conditions} \\
\midrule
0 & Unsafe & -- \\
1 & Minimal & $\text{IQM}(D_{\text{norm}}) \leq 0$ \\
2 & Moderate & $\text{IQM}(D_{\text{norm}}) \leq 0$ \; and \; $\text{IQM}(V) \leq 0.5$ \\
3 & Substantial & $\text{IQM}(D_{\text{norm}}) \leq 0$ \; and \; $\text{IQM}(V) \leq 0.1$ \; and \; $\text{IQM}(D_{\text{norm}^+}) \leq 0.1$ \\
4 & Strict & $\text{IQM}(D_{\text{norm}}) \leq 0$ \; and \; $\text{IQM}(V) = 0$ \; and \; $\text{IQM}(D_{\text{norm}^+}) = 0$ \\
\bottomrule
\end{tabular}
\label{tab:safety_tiers}
\end{table}

$\text{IQM}(D_{\text{norm}})\leq 0$ is required across all tiers from 1 upward, since it directly reflects an algorithm being safe on average. We therefore treat this as the minimal requirement for tier 1. However, $\text{IQM}(D_{\text{norm}})\leq 0$ alone does not guarantee infrequent violations. It is possible for the average cost to sit below the safety bound while the majority of episodes still violate it, with only large undershoots in a minority of episodes pushing the mean below the bound. We therefore impose an additional threshold for tier 2 at $\text{IQM}(V)\leq0.5$, such that the majority of episodes are not allowed to violate the safety bound. Because moving directly from tier 2 to strict safety would leave too large a gap between a moderately-safe algorithm and one that is fully constraint-satisfying, we introduce an intermediate tier 3, for which we set $\text{IQM}(V)\leq0.1$, and additionally set $\text{IQM}(D_{\text{norm}^+})\leq0.1$, so that, even under this low violation rate, the average overshoot on the episodes that do violate remains within $10\%$ of the safety bound. Tier 4 characterizes strict safety: the average cost is not allowed to exceed the safety bound, and no constraint violations are permitted at all. From $\text{IQM}(V)=0$, it then naturally follows that $\text{IQM}(D_{\text{norm}^+})$ is $0$ as well, since there are no violating episodes over which to compute an overshoot.

\paragraph{Aggregate cost distribution} \label{sec:CDF}
The metrics introduced in this section summarize an algorithm's safety only at the single threshold where cost equals the bound. To capture an algorithm's performance across the entire range of possible deviations from the safety bound, we recommend reporting the empirical cumulative distribution function (CDF) of the normalized episodic cost deviation. This provides an informative visualization of constraint adherence as a performance profile of an algorithm, and offers a visual way to compare algorithms across the entire set of episodes, rather than compressing this information into a single value, as recommended by \citet{agarwal2021deep}. For algorithm $a$, the aggregate empirical CDF evaluated at threshold $\kappa$ is given by Eq.~\ref{eq:CDF}.
\begin{align} \label{eq:CDF}
  \text{CDF}^{j}_a(\kappa) =
    \frac{\displaystyle\sum_{(m,d)\in\mathcal{P}}\sum_{k=1}^{K}\sum_{i=1}^{N}
          \mathbbm{1}\!\left[\tfrac{c_{i,(m,d,a,k)}-d}{d} \leq \kappa\right]}
         {|\mathcal{P}|\cdot K \cdot N},
\end{align}
where $c_{i,(m,d,a,k)}$ is the cost of the $i$-th episode, evaluated under policy $\pi_j$,
$N$ is the number of evaluated episodes per seed, and $\mathbbm{1}[\cdot]$ is the
indicator function. As with the other metrics introduced in this section, the CDF can be computed for a fixed policy $\pi_j$ at any training iterate $j$, for the final policy evaluated with exploration noise enabled or under a greedy policy, or aggregated across the entire course of training. This allows the effect of exploration noise on the full distribution of constraint adherence to be visualized directly, both over the course of training and at the final policy.

\section{Results} \label{sec: results}
For a detailed description of our experimental setup, we refer to Appendix Section \ref{sec: experimental setup}. The main focus of this paper is on the results of each metric aggregated over all tasks and bounds, shown in Table~\ref{tab:main_results}. Based on these aggregated metrics, we classify each algorithm into one of the safety tiers from Table~\ref{tab:safety_tiers}, both over the entire course of training and at the final greedy policy. We furthermore present results on the aggregated metrics evaluated on final policies with exploration noise on and off (greedy) in Table \ref{tab:main_results_expl_CI} in Appendix \ref{app: aggregate metrics}. Results for single tasks and bounds are reported in Appendix~\ref{app:per-condition results}, along with their training curves and episodic cost distributions. 


The results in Table~\ref{tab:main_results} show that all three safe RL algorithms reach a higher safety tier for a final greedy policy than over the entire course of training with exploration noise, which is expected for the selection of studied algorithms, since all of them only guarantee constraint satisfaction asymptotically or within a local trust region, rather than at every individual policy iterate encountered during training. These results also show that FOCOPS is marginally safer than PPO-Lag in terms of violation rate over training, yet FOCOPS is assigned tier 0 for training while PPO-Lag reaches tier 2. This is because FOCOPS's $\text{IQM}(D_{\text{norm}})$ over training is $0.06$, narrowly above the threshold of $\text{IQM}(D_{\text{norm}})\leq 0$ required for tier 1 and above. Nevertheless, FOCOPS achieves the highest $\text{IQM}(\bar{R})$ at both training and for a final greedy policy, while simultaneously achieving the lowest $\text{IQM}(D_{\text{norm}})$, $\text{IQM}(V)$, and $\text{IQM}(D_{\text{norm}^+})$ for a final greedy policy. PPO-Lag, by contrast, is the most conservative algorithm during training, with $\text{IQM}(D_{\text{norm}^+})=0.25$, but achieves a substantially lower $\text{IQM}(\bar{R})$ than FOCOPS.

\begin{table}[H]
\centering
\small
\caption{Safety and performance evaluation across algorithms, aggregated over all $K=30$ seeds for all $(m,d)\in\mathcal{P}$ pairs, IQM reported on each metric, for both over the course of all policy iterates (Train) and final \textit{greedy} policy (Final) settings. CIs for this table, computed via stratified bootstrapping, are reported in Appendix Table \ref{tab:main_results_CI}. \textbf{Bold text} indicates best result in that specific column, excluding PPO, as this is not a safe RL algorithm. Lower is better for each metric except for $\text{IQM}(\bar{R})$. PPO-Lag reaches tier 2 for training, even though its $\text{IQM}(V)$ is slightly higher than for FOCOPS, which sits in tier 0 due to $\text{IQM}(D_{\text{norm}})=0.06$ for training.}
\resizebox{\textwidth}{!}{%
\begin{tabular}{lcccccccccccc}
\toprule
\multirow{2}{*}{\textbf{Algorithm}} & \multicolumn{2}{c}{$\text{IQM}(\bar{R})$} & \multicolumn{2}{c}{$\text{IQM}(\bar{C})$} & \multicolumn{2}{c}{$\text{IQM}(D_{\text{norm}})$} & \multicolumn{2}{c}{$\text{IQM}(V)$} & \multicolumn{2}{c}{$\text{IQM}(D_{\text{norm}^+})$} & \multicolumn{2}{c}{Tier} \\
\cmidrule(lr){2-3} \cmidrule(lr){4-5} \cmidrule(lr){6-7} \cmidrule(lr){8-9} \cmidrule(lr){10-11} \cmidrule(lr){12-13}
 & Train & Final & Train & Final & Train & Final & Train & Final & Train & Final & Train & Final \\
 & & (greedy) & & (greedy) & & (greedy) & & (greedy) & & (greedy) & & (greedy) \\
\midrule
\textcolor{gray}{PPO}  & \textcolor{gray}{$115.63$} & \textcolor{gray}{$101.19$} & \textcolor{gray}{$145.98$} & \textcolor{gray}{$83.60$} & \textcolor{gray}{$4.77$} & \textcolor{gray}{$2.32$} & \textcolor{gray}{$1.00$} & \textcolor{gray}{$0.94$} & \textcolor{gray}{$4.79$} & \textcolor{gray}{$2.50$} & \textcolor{gray}{0} & \textcolor{gray}{0} \\
PPO-Lag  & $77.72$ & $55.95$ & $\bm{26.62}$ & $23.50$ & $\bm{0.00}$ & $-0.10$ & $0.43$ & $0.36$ & $\bm{0.25}$ & $0.55$ & 2 & 2 \\
P3O      & $90.41$ & $64.77$ & $45.69$ & $27.70$ & $0.55$ & $0.01$ & $0.70$ & $0.45$ & $0.81$ & $0.57$ & 0 & 2 \\
FOCOPS   & $\bm{108.76}$ & $\bm{74.26}$ & $30.39$ & $\bm{20.88}$ & $0.06$ & $\bm{-0.19}$ & $\bm{0.41}$ & $\bm{0.30}$ & $0.34$ & $\bm{0.50}$ & 0 & 2 \\
\bottomrule
\end{tabular}}
\label{tab:main_results}
\end{table}

\begin{figure}[H]
    \centering
    \includegraphics[width=1\linewidth]{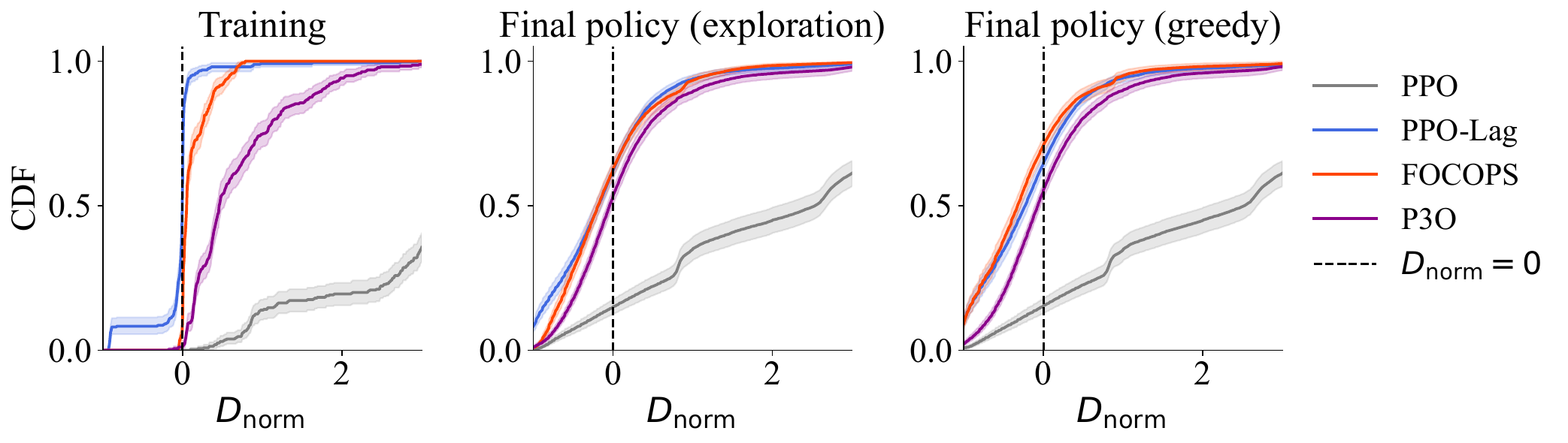}
    \caption{CDFs following Eq. \ref{eq:CDF}, over the entire range of training (left), for a final policy with exploration noise on (middle), and for a final greedy policy (right). Threshold $\kappa=0$ is indicated by the dashed vertical line at $D_{\text{norm}}=0$. Shaded regions indicate the 95\% CIs computed via stratified bootstrap. The $x$-axis is cut off at $D_{\text{norm}}=2$ for visual purposes. For a visualization across the entire range of the $x$-axis, see Figure~\ref{fig:agg_cdf_no_cutoff} in Appendix \ref{app: aggregate metrics}. Unlike the aggregate metrics in Table~\ref{tab:main_results}, the CDFs reveal the full shape of the underlying violation distribution across all thresholds, not just a single summary value. The CDF curves for the final policy with exploration noise are very similar to the curves of a final greedy policy, except for FOCOPS, where the final greedy policy (right) starts at roughly 15\% of episodes far below $D_{\text{norm}}$, while exploration noise removes this safe tail from the distribution (middle), starting at 0\%.}
    \label{fig:agg_cdf}
\end{figure}

Figure~\ref{fig:agg_cdf} shows the empirical CDFs computed by Eq.~\ref{eq:CDF}, for three settings: over the entire course of training, for the final policy evaluated with exploration noise enabled, and for the final, greedy policy. The ordering of the curves at $\kappa=0$ matches the ordering of the violation rate reported in Table~\ref{tab:main_results}. Note, however, that the values do not correspond one-to-one: Table~\ref{tab:main_results} reports $\text{IQM}(V)$, whereas the CDF is computed over all instances. The overall shape of each curve to the right of $\kappa=0$ reflects the severity of constraint violations across all conditions, rather than a single-value statistic. A curve that rises steeply immediately to the right of $\kappa=0$ indicates that, on the episodes where the safety bound is exceeded, overshoot tends to be small, whereas a curve with a longer tail extending further to the right instead reflects episodes with more severe violations, consistent with a higher $D_{\text{norm}^+}$.


The curves for all three safe RL algorithms shift visibly to the left when moving from training to the final policy, i.e., toward $\kappa=0$ and below, consistent with the improvement in safety tier from training to the final policy observed in Table~\ref{tab:main_results}. Notably, within the final policy, the curves for the policy with exploration noise are similar to the curves for the greedy policy, across all algorithms except for FOCOPS. For FOCOPS, the greedy policy has roughly 15\% of episodes with zero cost, while exploration noise erases this safe tail of the distribution. This is consistent with the results shown in Table~\ref{tab:main_results_expl_CI} in Appendix~\ref{app: aggregate metrics}, where exploration noise in the final policy accounts for a 9 percentage-point increase in violation rate and raises $\text{IQM}(D_{\text{norm}})$ by 0.142 relative to the greedy policy, with non-overlapping confidence intervals between the two evaluation modes. These results confirm that our proposed metrics and distributions reveal relevant information on safety that would otherwise be lost within a single cost metric: the confidence intervals of $\text{IQM}(\bar{C})$ for FOCOPS overlap between the two evaluation modes, whereas those of $\text{IQM}(D_{\text{norm}})$ and $\text{IQM}(V)$ do not.

\section{Discussion} \label{sec: Discussion}

While our proposed evaluation protocol provides a robust set of metrics that allow for a broadened view of safety and reliability, as well as aggregation across multiple tasks and bounds, we would like to point out that the classification into safety tiers can, in some cases, be sharply determined by a single condition. A small, consistent overshoot of bound $d$ can shift an algorithm by multiple tiers: for instance, in Table~\ref{tab:main_results}, FOCOPS's training $\text{IQM}(D_{\text{norm}})$ is only marginally above zero, yet this alone places it two tiers below PPO-Lag, despite FOCOPS achieving a comparable and even lower training $\text{IQM}(V)$. This illustrates that the thresholds introduced in Table \ref{tab:safety_tiers} can be sensitive to small differences near a boundary, and that the assigned safety tiers should be interpreted alongside the underlying metric values rather than as a standalone summary.

Additionally, aggregating results across tasks and safety bounds using the IQM is only meaningful for metrics that are directly comparable across tasks and safety bounds with potentially very different scales. $V$ is a rate bounded in $[0,1]$ by construction, and $D_{\text{norm}}$ and $D_{\text{norm}^+}$ are explicitly normalized by the safety bound $d$, making all three comparable across tasks and bounds. By contrast, $\bar{R}$ and $\bar{C}$ are not normalized and can vary substantially in scale across tasks, which is why our safety tiers are conditioned only on $V$, $D_{\text{norm}}$ and $D_{\text{norm}^+}$.

\subsection*{Limitations}
Our empirical evaluation is limited to four tasks and three safety bounds. While a broader evaluation across additional task domains \citep{tomilin_hasard_2025, ramanujam_safeor-gym_2025} and bounds would further strengthen the generality of our findings, the selected conditions were chosen to capture a broad range of diversity in constraint types and task complexity, providing a clear empirical basis for the results presented in this study, with approximately 2{,}880 GPU-hours on NVIDIA L4 GPUs with 24GB of memory. However, our results are limited to algorithms that permit constraint violations during training and only guarantee constraint adherence asymptotically or approximately. Stricter safe RL methods that disallow constraint violations even during training \citep{alshiekh2018safe, dalal2018safe, as2025actsafe, cheng2019end} are yet to be evaluated using our proposed framework. We therefore provide an open-source evaluation suite that enables the community to adopt our proposed evaluation metrics and apply them to other safe RL algorithms and task domains, which would help further extending the broadened view of safety and reliability that this work provides.


\subsection*{Future Work}
For safe real-world deployment, the gap between simulation-based and real-world deployment performance should ideally be negligible. Our results show small performance differences on a final policy with and without exploration noise for the algorithms evaluated in this work in simulation. Future work could compare the evaluation of simulated greedy policy rollouts against real-world deployment performance, using the distributions of $V$, $D_{\text{norm}}$, and $D_{\text{norm}^+}$ introduced here to quantify this gap directly.


We argue that a natural extension of this work is to incorporate the proposed evaluation metrics directly into algorithm design, rather than using them only for evaluation purposes. For example, an algorithm's training objective could explicitly minimize $V$, $D_{\text{norm}}$ or $D_{\text{norm}^+}$. This would allow safe RL algorithms to be optimized directly against the same safety and reliability criteria used to evaluate them.

\section{Conclusion} \label{sec: conclusion}
In this work, we introduced a comprehensive safe RL evaluation framework, including the violation rate ($V$), mean normalized cost deviation ($D_{\text{norm}}$), and normalized violation magnitude ($D_{\text{norm}^+}$), together with a safety tier system and a distributional view of constraint adherence, directly addressing the insufficiency of the expectation-based guarantee $\mathbb{E}[C]\leq d$ illustrated in Figure~\ref{fig:motivation}. Our evaluation study showed that no algorithm dominates uniformly across all conditions and both training and for a final policy: FOCOPS achieves the strongest overall balance of reward and safety at a final greedy policy, while PPO-Lag trades off reward for a lower violation magnitude over the course of training. We further found that the final greedy policy and the final policy with exploration noise yield broadly similar results on safety metrics across all evaluated algorithms, except for FOCOPS, where exploration noise in the final policy accounts for a significant increase in violation rate and $\text{IQM}(D_{\text{norm}})$, a discrepancy not captured by the average cost alone. Additionally, our results show that aggregate safety tiers, the CDF, and results per task and safety bound each reveal information that is crucial to a complete evaluation of an algorithm's safety and reliability. We therefore recommend reporting all three jointly, rather than mean cost or mean reward alone, and provide an open-source evaluation suite to provide the tools to assess safe RL algorithms by their reliability, not just their average performance.




\begin{ack}
This work was supported by Shell Information Technology International Limited and the Netherlands Enterprise Agency under the grant PPS23-3-03529461.
\end{ack}

\bibliography{EvaluateSafeRL}
\bibliographystyle{ACM-Reference-Format}

\newpage
\appendix
\renewcommand{\thesection}{A.\arabic{section}}
\renewcommand{\thesubsection}{A.\arabic{section}.\arabic{subsection}}
\renewcommand{\thefigure}{A.\arabic{figure}}
\renewcommand{\thetable}{A.\arabic{table}}
\setcounter{section}{0}
\setcounter{figure}{0}
\setcounter{table}{0}

\section{Experimental setup} \label{sec: experimental setup}

We train algorithms PPO-Lag \citep{ray_benchmarking_nodate}, P3O \citep{zhang2022penalized}, FOCOPS \citep{zhang_first_2020} and, as an unconstrained baseline, PPO \citep{schulman_proximal_2017} using the CRAX benchmark suite \citep{tomilin2026crax}, for 30 seeds each. The set of tasks is $\mathcal{M}=\{\text{Safe Goal Point, Safe Circle Point, Safe Push Point, Safe Button Point}\}$, all Level 1, and the set of safety bounds is $\mathcal{D}=\{15,25,50\}$. This set of tasks focuses on safe navigation and consists of different types of constraints and rewards, selected carefully to span a set of tasks with sufficient variety in their cost functions. 


We train each run for $5\cdot10^8$ total environment steps, with an episode length of $T=2000$ steps. During training, actions are sampled stochastically from the learned Gaussian policy distribution (exploration noise on), and training metrics are recorded once per complete batch: after every cycle in which all $N=2048$ parallel environments finish their episode (every ${\sim}4.2\cdot10^6$ environment steps), yielding $J=122$ data points per seed over $5\cdot10^8$ total environment steps. Each data point is averaged over the $2048$ episodes in that batch, for a total of approximately $2.5\cdot10^5$ episodes per seed. At the final policy evaluation we report two evaluations over $N=100$ episodes per seed: a stochastic evaluation that samples actions from the policy distribution (matching training conditions), and a deterministic evaluation that takes the distribution mean (greedy policy, no exploration noise). 

 For metric $X\in\{\bar{R},\bar{C},D_{\text{norm}}, V, D^+_{\text{norm}}\}$, we report the task- and bound-specific results as $\frac{1}{K}\sum_{k=1}^{K} X_{(m,d,a,k)} \pm \sigma_k$, where $\sigma_k$ is the sample standard deviation for seed $k$, for $K=30$ seeds. This standard deviation reflects the variability across individual training runs under identical conditions, and does not shrink with additional seeds. For aggregation over all tasks and safety bounds, we use the IQM as explained in Section \ref{sec:Aggregation}. Since the IQM is a nonlinear statistic of the distribution across conditions, we report uncertainty as a 95\% confidence interval (CI) computed via a stratified bootstrap with $B=2000$ replicates \citep{efron1992bootstrap, agarwal2021deep}. Unlike the standard deviation in a single task or bound, this confidence interval reflects the standard error of the IQM estimate itself, i.e., how much the aggregated IQM value would be expected to change under a different sample of seeds, and would narrow with additional seeds or conditions. We therefore report both: the standard deviation characterizes the inherent variability of individual training runs, while the IQM confidence interval characterizes the reliability of our aggregate statistics. Likewise, the CDF curve following Eq.~\ref{eq:CDF} is estimated from finite data ($K$ seeds $\times$ $N$ episodes per condition), so we used the same stratified bootstrap procedure to report the CI band showing how much the curve would shift under different seeds.

\paragraph{Compute resources} All experiments were run on NVIDIA L4 GPUs (24\,GB memory), using a single GPU and 4 CPU cores per run. A single training run of 5$\cdot$10$^8$ environment steps took approximately 2 hours of wall-clock time. Our full experimental suite consists of 4 tasks $\times$ 3 safety bounds $\times$ 4 algorithms (PPO, PPO-Lag, P3O, FOCOPS) $\times$ 30 seeds, totaling 1{,}440 runs, for an estimated total of approximately 2{,}880 GPU-hours, or equivalently, approximately 11{,}520 CPU-core-hours, given 4 cores allocated per run.

\newpage
\begin{table}[t]
\centering
\footnotesize
\caption{Hyperparameters used across all experiments.}
\begin{tabularx}{\textwidth}{Xc Xc}
\toprule
\multicolumn{4}{l}{\textit{Shared (all algorithms)}} \\
\cmidrule(r){1-2}\cmidrule(l){3-4}
\textbf{Hyperparameter} & \textbf{Value} & \textbf{Hyperparameter} & \textbf{Value} \\
\midrule
Total environment steps        & $5 \times 10^8$     & Discount factor $\gamma$       & $0.99$ \\
Episode length                 & $2000$              & GAE $\lambda$                  & $0.95$ \\
Parallel environments          & $2048$              & PPO clipping $\varepsilon$     & $0.3$ \\
Unroll length                  & $8$                 & Entropy coefficient            & $5 \times 10^{-3}$ \\
Batch size                     & $1024$              & Reward scaling                 & $0.1$ \\
Number of minibatches          & $32$                & Observation normalisation      & \checkmark \\
SGD epochs per batch           & $6$                 & Deterministic eval.\ episodes  & $100$ \\
Learning rate                  & $5 \times 10^{-4}$ & Seeds                          & $30$ \\
\midrule
\multicolumn{4}{l}{\textit{Algorithm-specific}} \\
\cmidrule(r){1-2}\cmidrule(l){3-4}
\multicolumn{2}{l}{\textit{PPO-Lag}} & \multicolumn{2}{l}{\textit{FOCOPS}} \\
\midrule
Lagrange multiplier LR         & $3.0$               & Initial $\nu$                  & $0.1$ \\
Initial $\lambda$              & $0.0$               & $\nu$ learning rate            & $1.0$ \\
                               &                     & $\nu_{\max}$                   & $100.0$ \\
                               &                     & KL penalty $\lambda$           & $1.5$ \\
                               &                     & Advantage temperature $\eta$   & $0.02$ \\
\cmidrule(r){1-2}
\multicolumn{2}{l}{\textit{P3O}} & & \\
\midrule
Initial $\kappa$               & $0.01$              & & \\
$\kappa$ increase factor       & $1.1$               & & \\
$\kappa_{\max}$                & $50.0$              & & \\
\bottomrule
\end{tabularx}
\label{tab:hyperparameters}
\end{table}

\newpage
\section{Additional results}

\subsection{Aggregate metrics} \label{app: aggregate metrics}

\begin{figure}[H]
    \centering
    \includegraphics[width=0.8\linewidth]{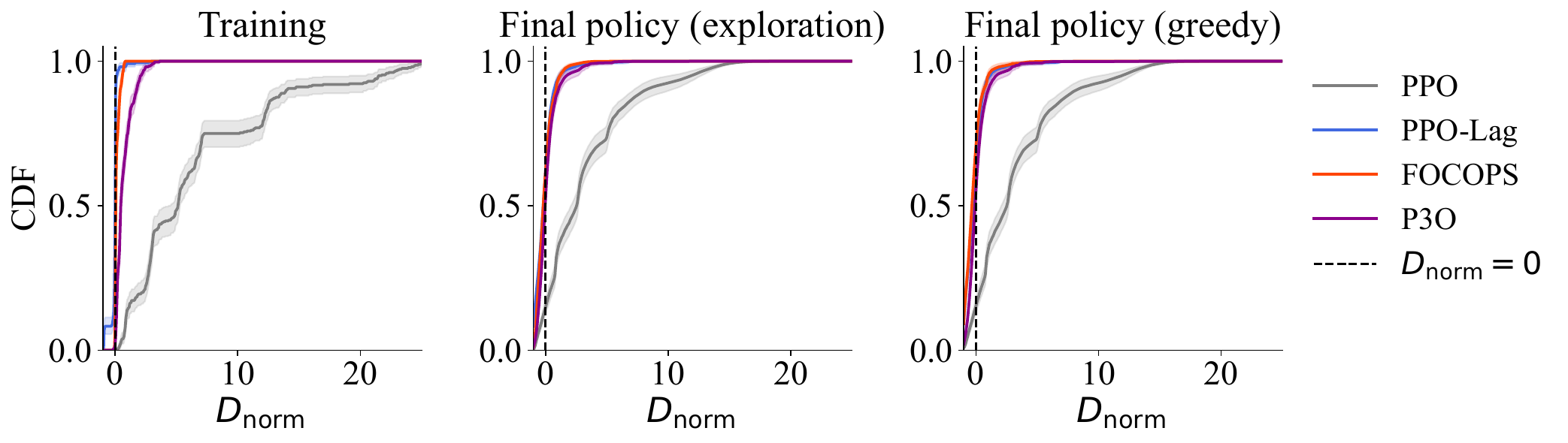}
    \caption{CDFs following Eq. \ref{eq:CDF}, over the entire range of training (left), for a final policy with exploration noise on (middle), and for a final greedy policy (right). Threshold $\kappa=0$ is indicated by the dashed vertical line at $D_{\text{norm}}=0$. Shaded regions indicate the 95\% CIs computed via stratified bootstrap.}
    \label{fig:agg_cdf_no_cutoff}
\end{figure}

Figure \ref{fig:agg_cdf_no_cutoff} shows the empirical CDFs computed by Eq. \ref{eq:CDF}, aggregated over all tasks, safety bounds, and seeds. Table \ref{tab:main_results_expl_CI} presents algorithm performance on the aggregated metrics evaluated on final policies with exploration noise on (expl.) and off (greedy), together with their stratified bootstrapped CIs. 

\begin{sidewaystable}[p]
\centering
\caption{Safety and performance evaluation across algorithms, aggregated over all $(m,d)$ pairs. IQM reported with 95\% stratified bootstrap CI $[\text{lower},\text{upper}]$ ($B=2000$), comparing the final policy evaluated with exploration noise enabled against the final, greedy policy. \textbf{Bold} indicates best result per column, excluding PPO.}
\resizebox{\textwidth}{!}{%
\begin{tabular}{lcccccccccccc}
\toprule
\multirow{2}{*}{\textbf{Algorithm}} & \multicolumn{2}{c}{$\text{IQM}(\bar{R})$} & \multicolumn{2}{c}{$\text{IQM}(\bar{C})$} & \multicolumn{2}{c}{$\text{IQM}(D_{\text{norm}})$} & \multicolumn{2}{c}{$\text{IQM}(V)$} & \multicolumn{2}{c}{$\text{IQM}(D_{\text{norm}}^{+})$} & \multicolumn{2}{c}{Tier} \\
\cmidrule(lr){2-3} \cmidrule(lr){4-5} \cmidrule(lr){6-7} \cmidrule(lr){8-9} \cmidrule(lr){10-11} \cmidrule(lr){12-13}
 & Final (expl.) & Final (greedy) & Final (expl.) & Final (greedy) & Final (expl.) & Final (greedy) & Final (expl.) & Final (greedy) & Final (expl.) & Final (greedy) & Final (expl.) & Final (greedy) \\
\midrule
\textcolor{gray}{PPO}
  & \textcolor{gray}{$101.675_{\scriptscriptstyle[100.121,103.147]}$}
  & \textcolor{gray}{$101.193_{\scriptscriptstyle[99.500,102.639]}$}
  & \textcolor{gray}{$84.464_{\scriptscriptstyle[82.324,86.531]}$}
  & \textcolor{gray}{$83.597_{\scriptscriptstyle[81.311,85.830]}$}
  & \textcolor{gray}{$2.345_{\scriptscriptstyle[2.258,2.428]}$}
  & \textcolor{gray}{$2.318_{\scriptscriptstyle[2.232,2.403]}$}
  & \textcolor{gray}{$0.949_{\scriptscriptstyle[0.932,0.964]}$}
  & \textcolor{gray}{$0.941_{\scriptscriptstyle[0.922,0.957]}$}
  & \textcolor{gray}{$2.508_{\scriptscriptstyle[2.433,2.587]}$}
  & \textcolor{gray}{$2.500_{\scriptscriptstyle[2.420,2.573]}$}
  & \textcolor{gray}{0} & \textcolor{gray}{0} \\
PPO-Lag
  & $56.645_{\scriptscriptstyle[54.483,59.087]}$
  & $55.950_{\scriptscriptstyle[53.577,58.474]}$
  & $\bm{24.064}_{\scriptscriptstyle[22.204,26.590]}$
  & $23.502_{\scriptscriptstyle[21.630,26.148]}$
  & $\bm{-0.058}_{\scriptscriptstyle[-0.095,-0.027]}$
  & $-0.095_{\scriptscriptstyle[-0.164,-0.054]}$
  & $0.396_{\scriptscriptstyle[0.377,0.420]}$
  & $0.364_{\scriptscriptstyle[0.335,0.395]}$
  & $\bm{0.481}_{\scriptscriptstyle[0.450,0.532]}$
  & $0.549_{\scriptscriptstyle[0.487,0.613]}$
  & 2 & 2 \\
P3O
  & $65.446_{\scriptscriptstyle[63.606,67.861]}$
  & $64.770_{\scriptscriptstyle[63.033,67.392]}$
  & $28.643_{\scriptscriptstyle[26.555,30.972]}$
  & $27.700_{\scriptscriptstyle[25.845,29.990]}$
  & $0.029_{\scriptscriptstyle[-0.001,0.072]}$
  & $0.013_{\scriptscriptstyle[-0.020,0.056]}$
  & $0.456_{\scriptscriptstyle[0.434,0.476]}$
  & $0.449_{\scriptscriptstyle[0.423,0.464]}$
  & $0.539_{\scriptscriptstyle[0.498,0.580]}$
  & $0.574_{\scriptscriptstyle[0.524,0.611]}$
  & 2 & 2 \\
FOCOPS
  & $\bm{70.176}_{\scriptscriptstyle[68.728,71.685]}$
  & $\bm{74.257}_{\scriptscriptstyle[72.672,75.784]}$
  & $24.619_{\scriptscriptstyle[23.082,26.022]}$
  & $\bm{20.881}_{\scriptscriptstyle[19.173,23.209]}$
  & $-0.051_{\scriptscriptstyle[-0.098,-0.001]}$
  & $\bm{-0.193}_{\scriptscriptstyle[-0.243,-0.123]}$
  & $\bm{0.388}_{\scriptscriptstyle[0.357,0.420]}$
  & $\bm{0.297}_{\scriptscriptstyle[0.265,0.325]}$
  & $0.486_{\scriptscriptstyle[0.423,0.537]}$
  & $\bm{0.499}_{\scriptscriptstyle[0.439,0.547]}$
  & 2 & 2 \\
\bottomrule
\end{tabular}}
\label{tab:main_results_expl_CI}

\vspace{5em}

\caption{Safety and performance evaluation across algorithms, aggregated over all $(m,d)$ pairs. IQM reported with 95\% stratified bootstrap CI $[\text{lower},\text{upper}]$ ($B=2000$), for both over the course of all policy iterates (Train) and final \textit{greedy} policy (Final) settings. \textbf{Bold} indicates best result per column, excluding PPO.}
\resizebox{\textwidth}{!}{%
\begin{tabular}{lcccccccccccc}
\toprule
\multirow{2}{*}{\textbf{Algorithm}} & \multicolumn{2}{c}{$\text{IQM}(\bar{R})$} & \multicolumn{2}{c}{$\text{IQM}(\bar{C})$} & \multicolumn{2}{c}{$\text{IQM}(D_{\text{norm}})$} & \multicolumn{2}{c}{$\text{IQM}(V)$} & \multicolumn{2}{c}{$\text{IQM}(D_{\text{norm}^+})$} & \multicolumn{2}{c}{Tier} \\
\cmidrule(lr){2-3} \cmidrule(lr){4-5} \cmidrule(lr){6-7} \cmidrule(lr){8-9} \cmidrule(lr){10-11} \cmidrule(lr){12-13}
 & Train & Final (greedy) & Train & Final (greedy) & Train & Final (greedy) & Train & Final (greedy) & Train & Final (greedy) & Train & Final (greedy) \\
\midrule
\textcolor{gray}{PPO}
  & \textcolor{gray}{$115.63_{\scriptscriptstyle[114.87,116.38]}$}
  & \textcolor{gray}{$101.19_{\scriptscriptstyle[99.50,102.64]}$}
  & \textcolor{gray}{$145.98_{\scriptscriptstyle[145.18,146.83]}$}
  & \textcolor{gray}{$83.60_{\scriptscriptstyle[81.31,85.83]}$}
  & \textcolor{gray}{$4.77_{\scriptscriptstyle[4.70,4.85]}$}
  & \textcolor{gray}{$2.32_{\scriptscriptstyle[2.23,2.40]}$}
  & \textcolor{gray}{$1.00_{\scriptscriptstyle[1.00,1.00]}$}
  & \textcolor{gray}{$0.94_{\scriptscriptstyle[0.92,0.96]}$}
  & \textcolor{gray}{$4.79_{\scriptscriptstyle[4.72,4.87]}$}
  & \textcolor{gray}{$2.50_{\scriptscriptstyle[2.42,2.57]}$}
  & \textcolor{gray}{0} & \textcolor{gray}{0} \\
PPO-Lag
  & $77.72_{\scriptscriptstyle[73.31,80.50]}$
  & $55.95_{\scriptscriptstyle[53.58,58.47]}$
  & $\bm{26.62}_{\scriptscriptstyle[25.69,27.77]}$
  & $23.50_{\scriptscriptstyle[21.63,26.15]}$
  & $\bm{0.00}_{\scriptscriptstyle[-0.01,0.00]}$
  & $-0.10_{\scriptscriptstyle[-0.16,-0.05]}$
  & $0.43_{\scriptscriptstyle[0.42,0.44]}$
  & $0.36_{\scriptscriptstyle[0.34,0.40]}$
  & $\bm{0.25}_{\scriptscriptstyle[0.23,0.28]}$
  & $0.55_{\scriptscriptstyle[0.49,0.61]}$
  & 2 & 2 \\
P3O
  & $90.41_{\scriptscriptstyle[89.32,91.55]}$
  & $64.77_{\scriptscriptstyle[63.03,67.39]}$
  & $45.69_{\scriptscriptstyle[44.70,46.75]}$
  & $27.70_{\scriptscriptstyle[25.85,29.99]}$
  & $0.55_{\scriptscriptstyle[0.52,0.57]}$
  & $0.01_{\scriptscriptstyle[-0.02,0.06]}$
  & $0.70_{\scriptscriptstyle[0.69,0.72]}$
  & $0.45_{\scriptscriptstyle[0.42,0.46]}$
  & $0.81_{\scriptscriptstyle[0.78,0.83]}$
  & $0.57_{\scriptscriptstyle[0.52,0.61]}$
  & 0 & 0 \\
FOCOPS
  & $\bm{108.76}_{\scriptscriptstyle[107.77,109.78]}$
  & $\bm{74.26}_{\scriptscriptstyle[72.67,75.78]}$
  & $30.39_{\scriptscriptstyle[30.26,30.52]}$
  & $\bm{20.88}_{\scriptscriptstyle[19.17,23.21]}$
  & $0.06_{\scriptscriptstyle[0.06,0.06]}$
  & $\bm{-0.19}_{\scriptscriptstyle[-0.24,-0.12]}$
  & $\bm{0.41}_{\scriptscriptstyle[0.40,0.41]}$
  & $\bm{0.30}_{\scriptscriptstyle[0.27,0.33]}$
  & $0.34_{\scriptscriptstyle[0.33,0.34]}$
  & $\bm{0.50}_{\scriptscriptstyle[0.44,0.55]}$
  & 0 & 2 \\
\bottomrule
\end{tabular}}
\label{tab:main_results_CI}
\end{sidewaystable}

\newpage
\subsection{Task- and bound-specific results} \label{app:per-condition results}
\begin{figure}[H]
    \centering
    \begin{subfigure}[b]{0.32\textwidth}
        \centering
        \includegraphics[width=1.1\textwidth]{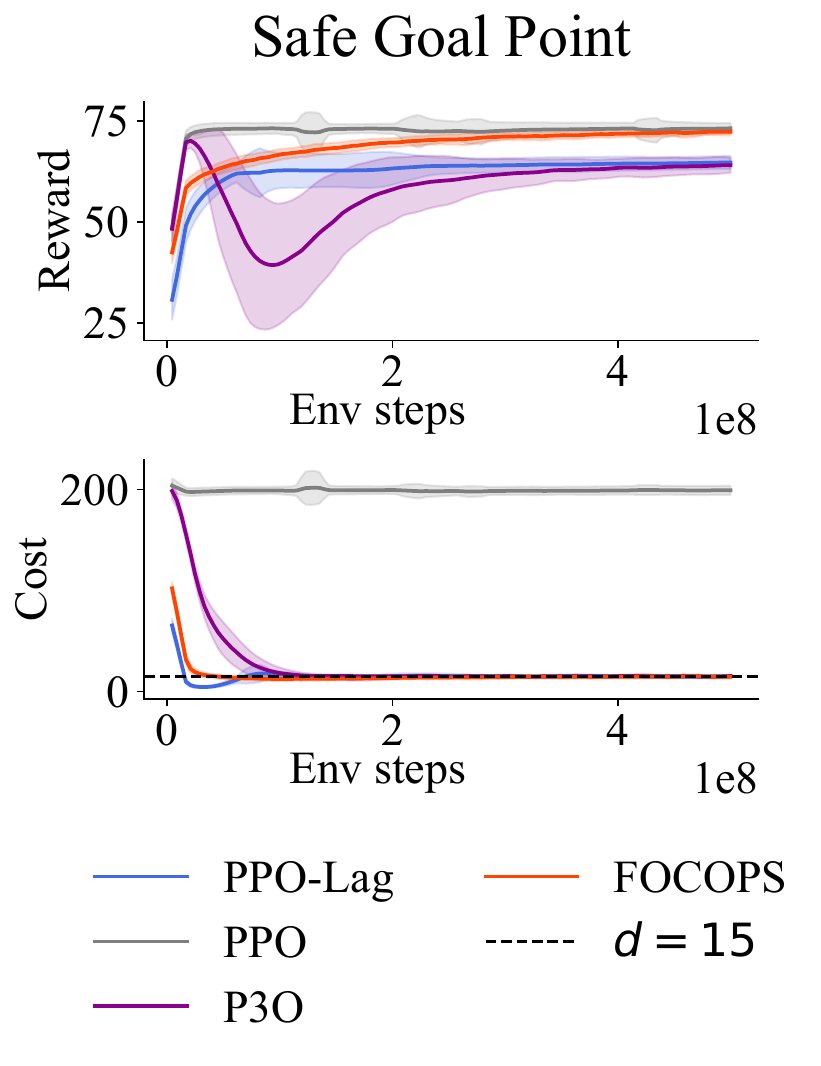}
        \caption{$d=15$}
    \end{subfigure}
    \hfill
    \begin{subfigure}[b]{0.32\textwidth}
        \centering
        \includegraphics[width=1.1\textwidth]{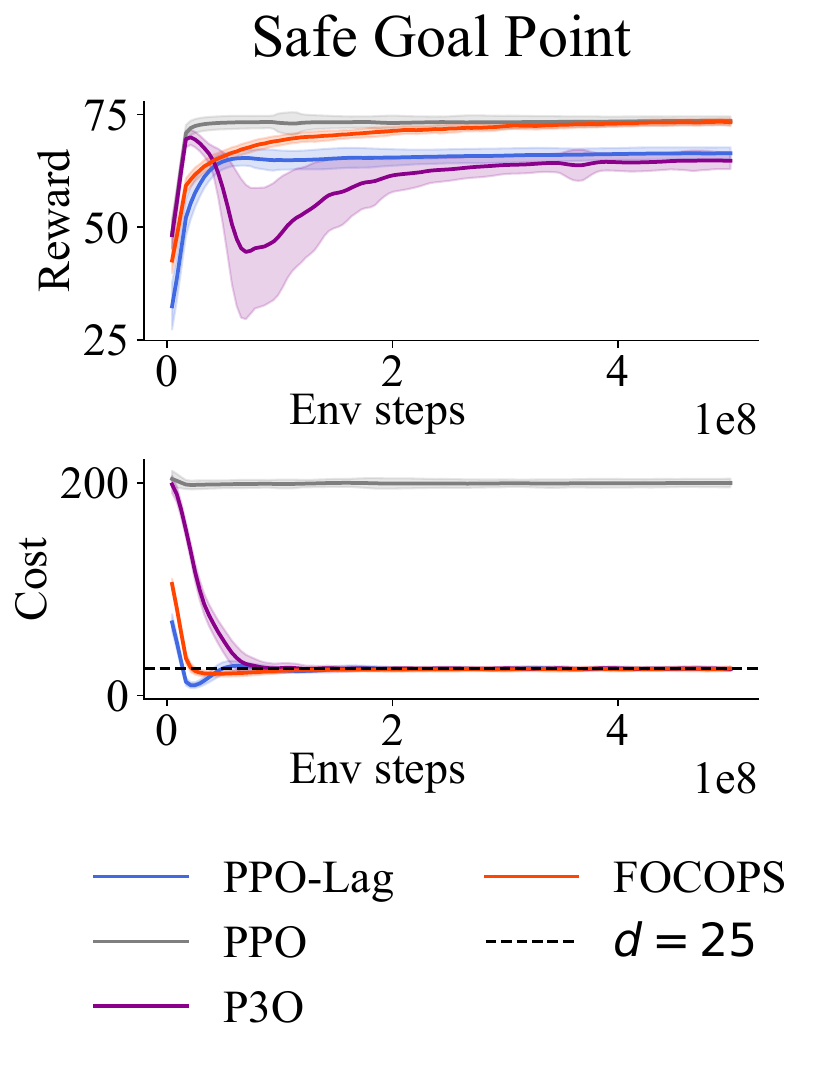}
        \caption{$d=25$}
    \end{subfigure}
    \hfill
    \begin{subfigure}[b]{0.32\textwidth}
        \centering
        \includegraphics[width=1.1\textwidth]{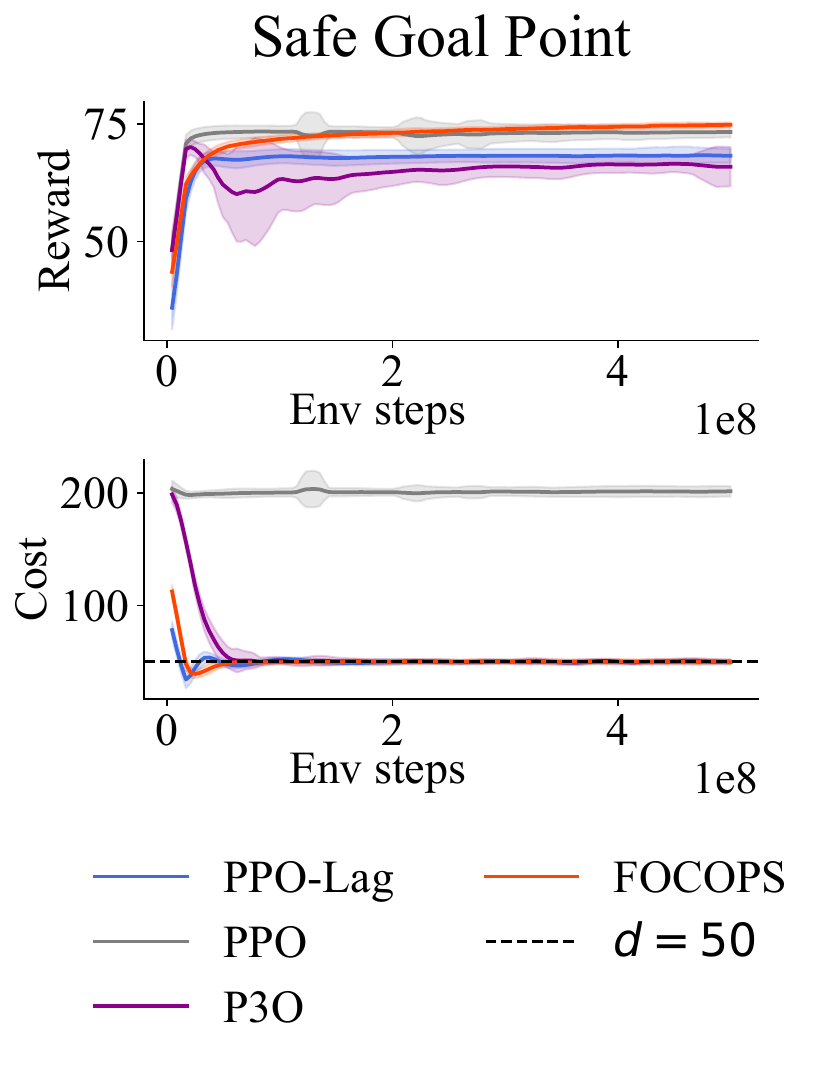}
        \caption{$d=50$}
    \end{subfigure}
    \caption{Smoothed training curves for Safe Goal Point (Level 1), for safety bounds $d\in\{15,25,50\}$, averaged over 30 seeds. The shaded regions denote the $1\sigma$ over 30 seeds.}
    \label{fig:seed_curves_goal}
\end{figure}

\begin{figure}[H]
    \centering
    \begin{subfigure}[b]{0.32\textwidth}
        \centering
        \includegraphics[width=1.1\textwidth]{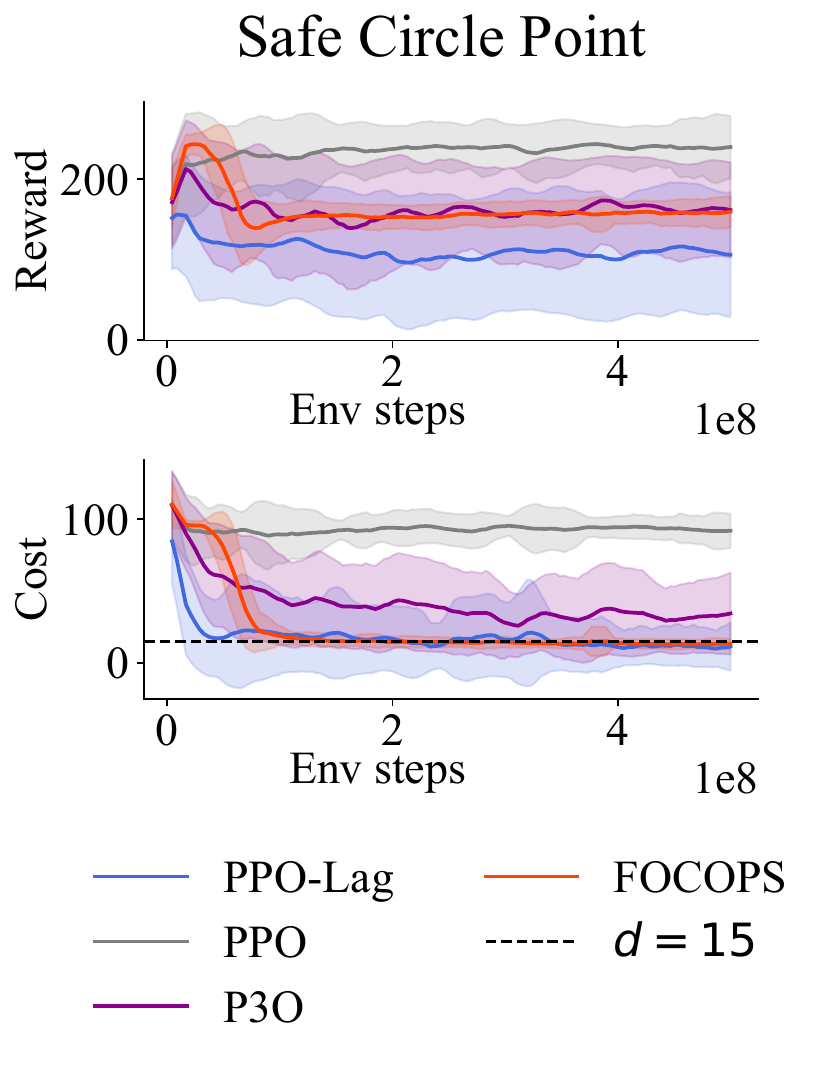}
        \caption{$d=15$}
    \end{subfigure}
    \hfill
    \begin{subfigure}[b]{0.32\textwidth}
        \centering
        \includegraphics[width=1.1\textwidth]{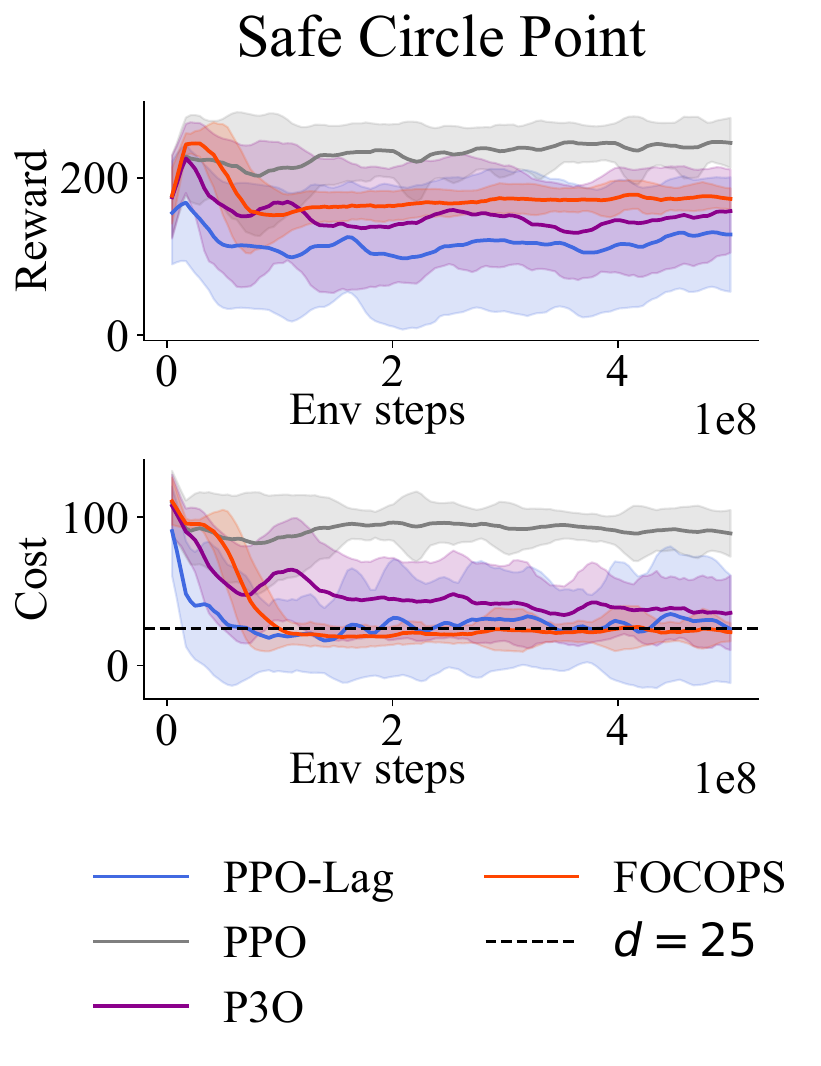}
        \caption{$d=25$}
    \end{subfigure}
    \hfill
    \begin{subfigure}[b]{0.32\textwidth}
        \centering
        \includegraphics[width=1.1\textwidth]{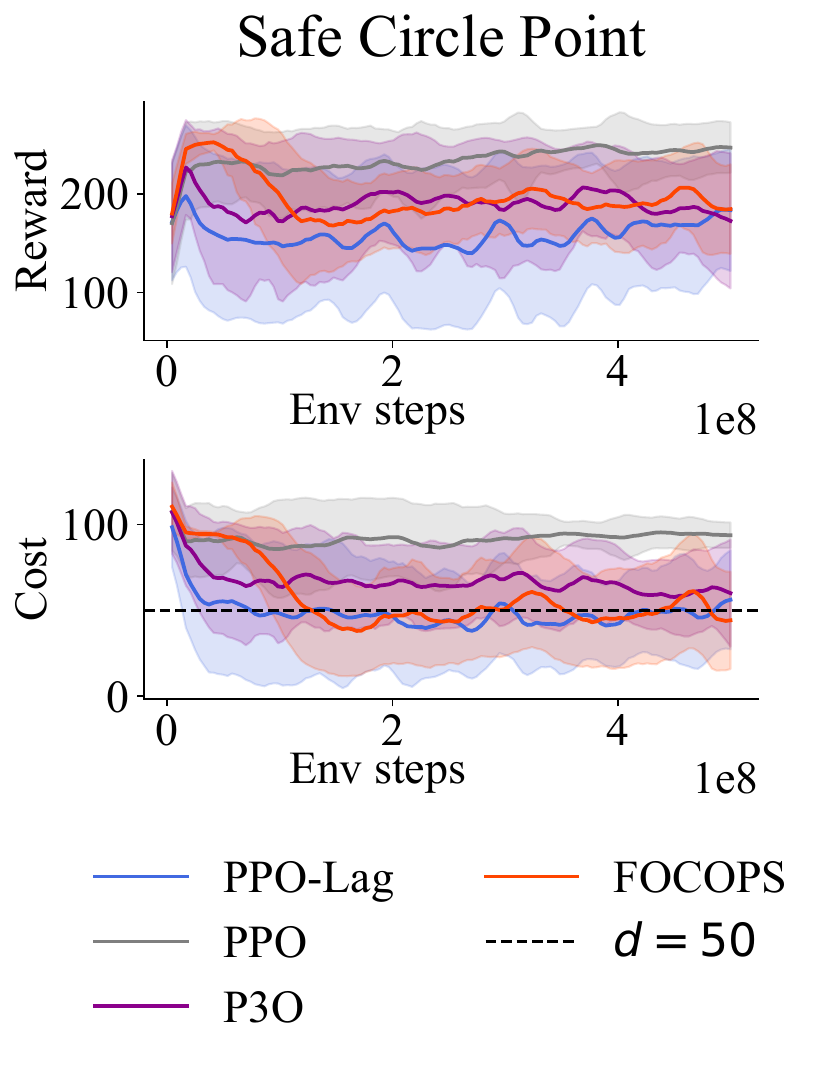}
        \caption{$d=50$}
    \end{subfigure}
    \caption{Smoothed training curves for Safe Circle Point (Level 1), for safety bounds $d\in\{15,25,50\}$, averaged over 30 seeds. The shaded regions denote the $1\sigma$ over 30 seeds.}
    \label{fig:seed_curves_circle}
\end{figure}

\begin{figure}[H]
    \centering
    \begin{subfigure}[b]{0.32\textwidth}
        \centering
        \includegraphics[width=1.1\textwidth]{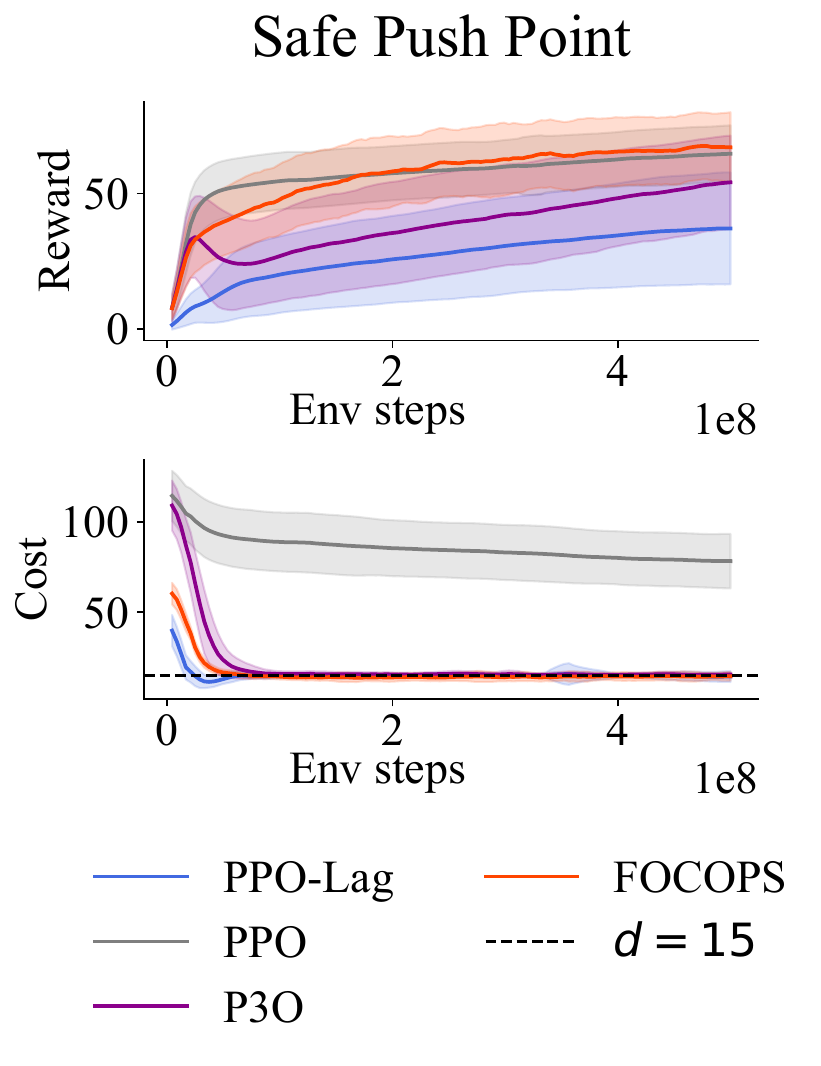}
        \caption{$d=15$}
    \end{subfigure}
    \hfill
    \begin{subfigure}[b]{0.32\textwidth}
        \centering
        \includegraphics[width=1.1\textwidth]{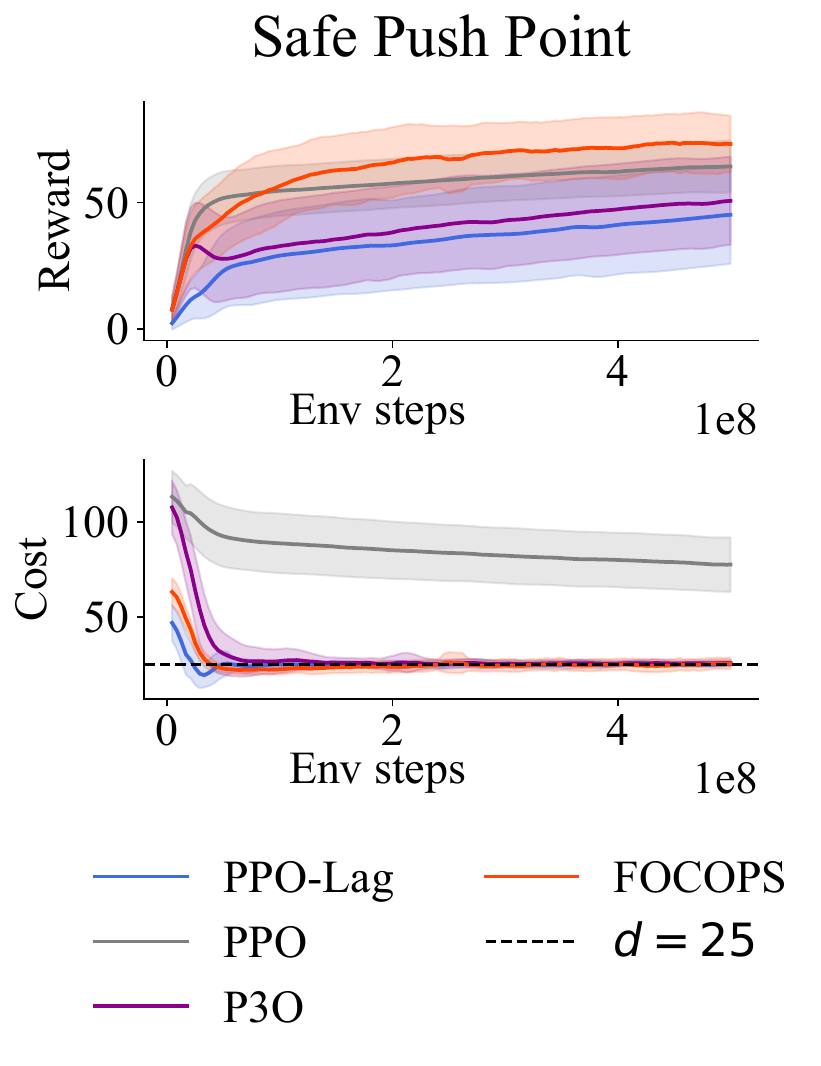}
        \caption{$d=25$}
    \end{subfigure}
    \hfill
    \begin{subfigure}[b]{0.32\textwidth}
        \centering
        \includegraphics[width=1.1\textwidth]{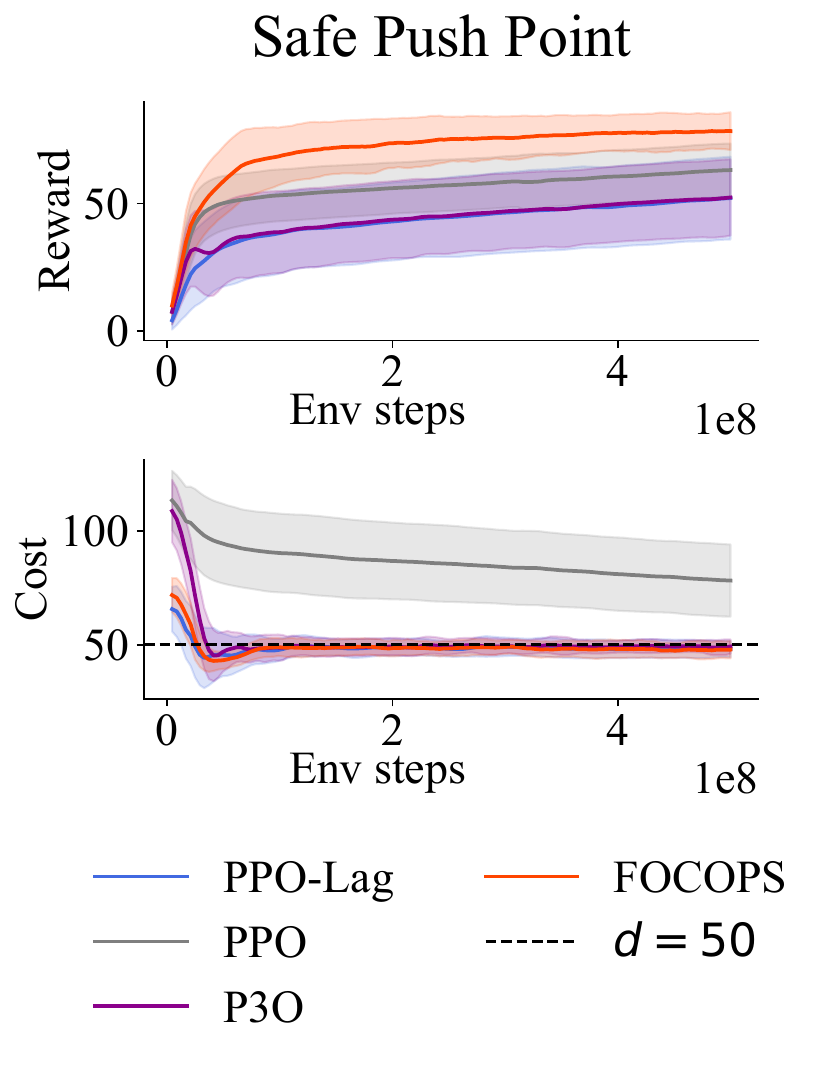}
        \caption{$d=50$}
    \end{subfigure}
    \caption{Smoothed training curves for Safe Push Point (Level 1), for safety bounds $d\in\{15,25,50\}$, averaged over 30 seeds. The shaded regions denote the $1\sigma$ over 30 seeds.}
    \label{fig:seed_curves_push}
\end{figure}

\begin{figure}[H]
    \centering
    \begin{subfigure}[b]{0.32\textwidth}
        \centering
        \includegraphics[width=1.1\textwidth]{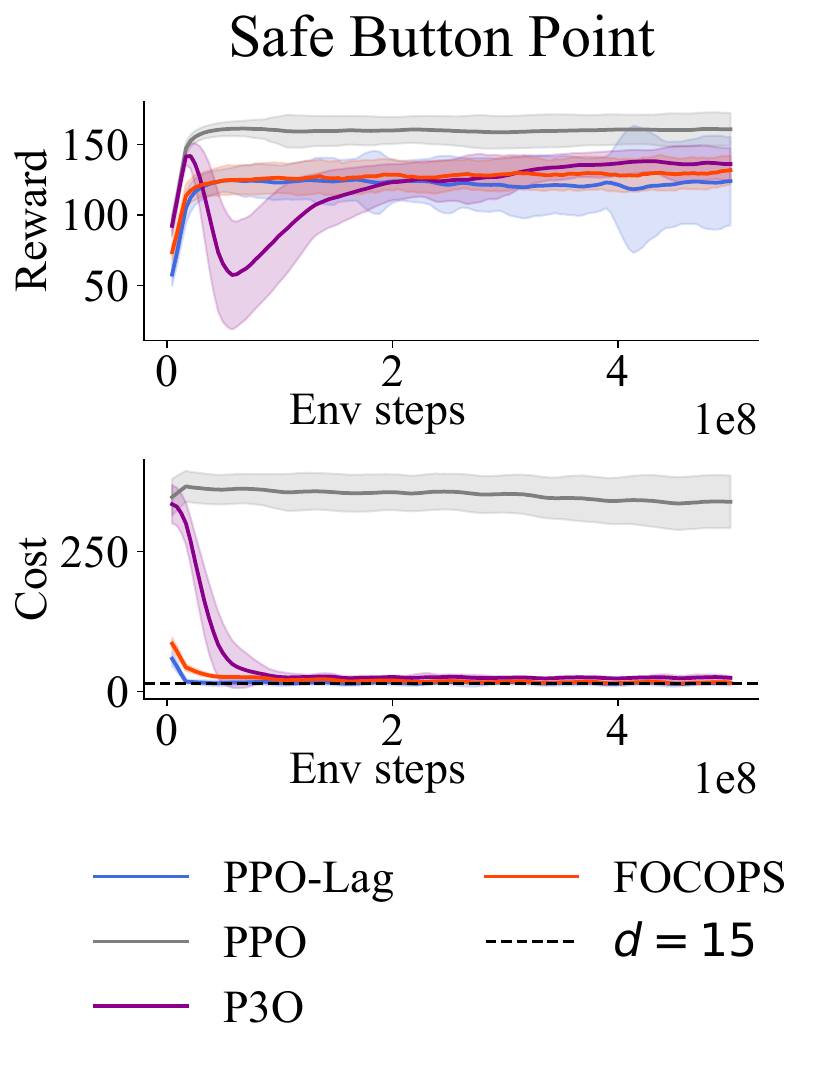}
        \caption{$d=15$}
    \end{subfigure}
    \hfill
    \begin{subfigure}[b]{0.32\textwidth}
        \centering
        \includegraphics[width=1.1\textwidth]{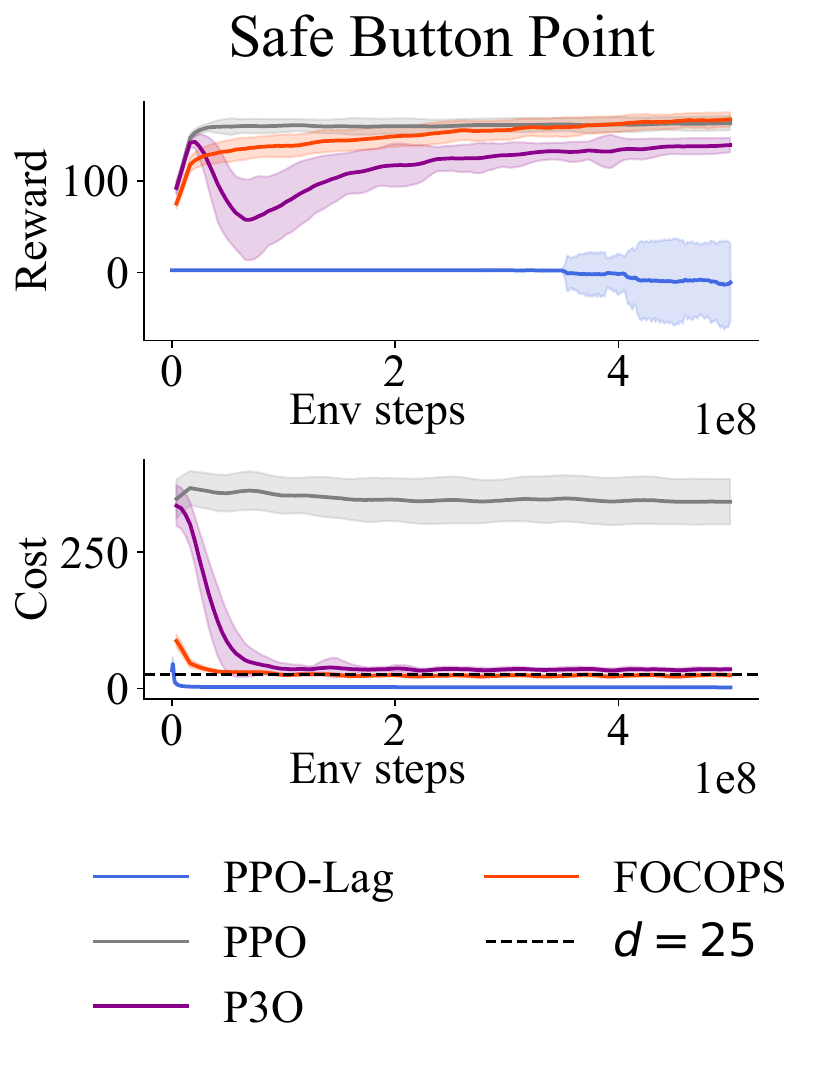}
        \caption{$d=25$}
        \label{fig:seed_curves_button_25}
    \end{subfigure}
    \hfill
    \begin{subfigure}[b]{0.32\textwidth}
        \centering
        \includegraphics[width=1.1\textwidth]{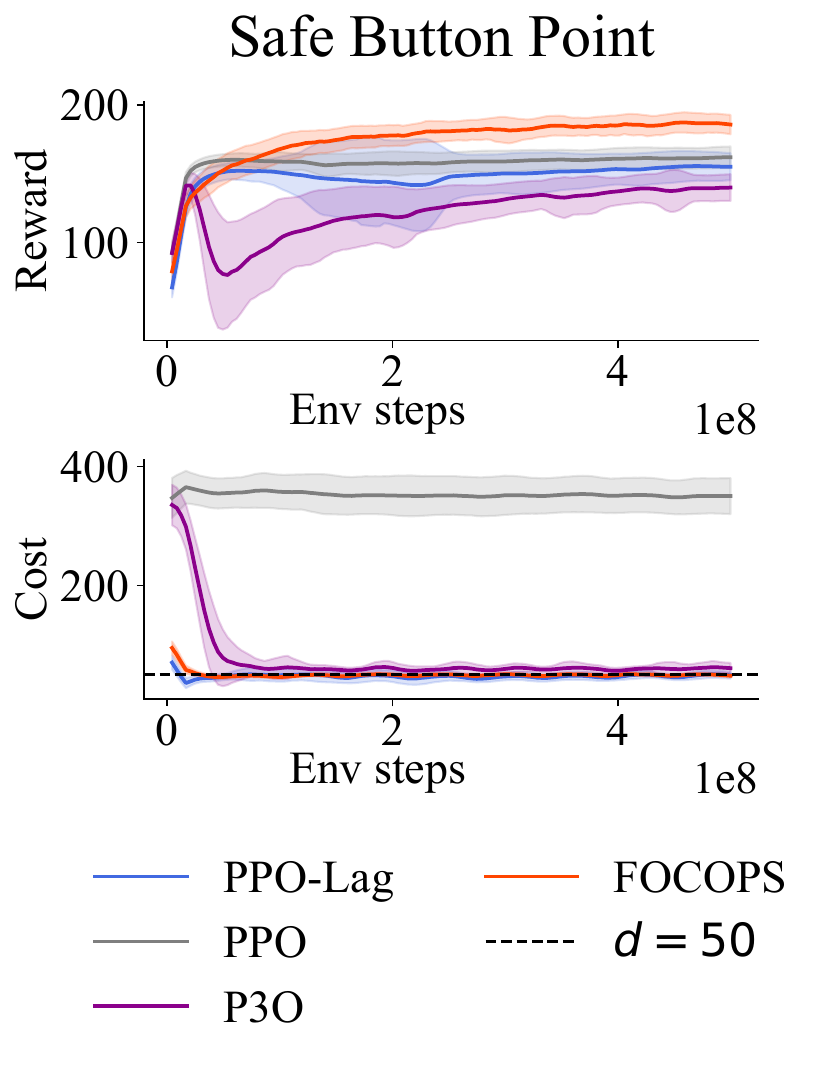}
        \caption{$d=50$}
    \end{subfigure}
    \caption{Smoothed training curves for Safe Button Point (Level 1), for safety bounds $d\in\{15,25,50\}$, averaged over 30 seeds. The shaded regions denote the $1\sigma$ over 30 seeds.}
    \label{fig:seed_curves_button}
\end{figure}

\begin{figure}[H]
    \centering
    \includegraphics[width=1\linewidth]{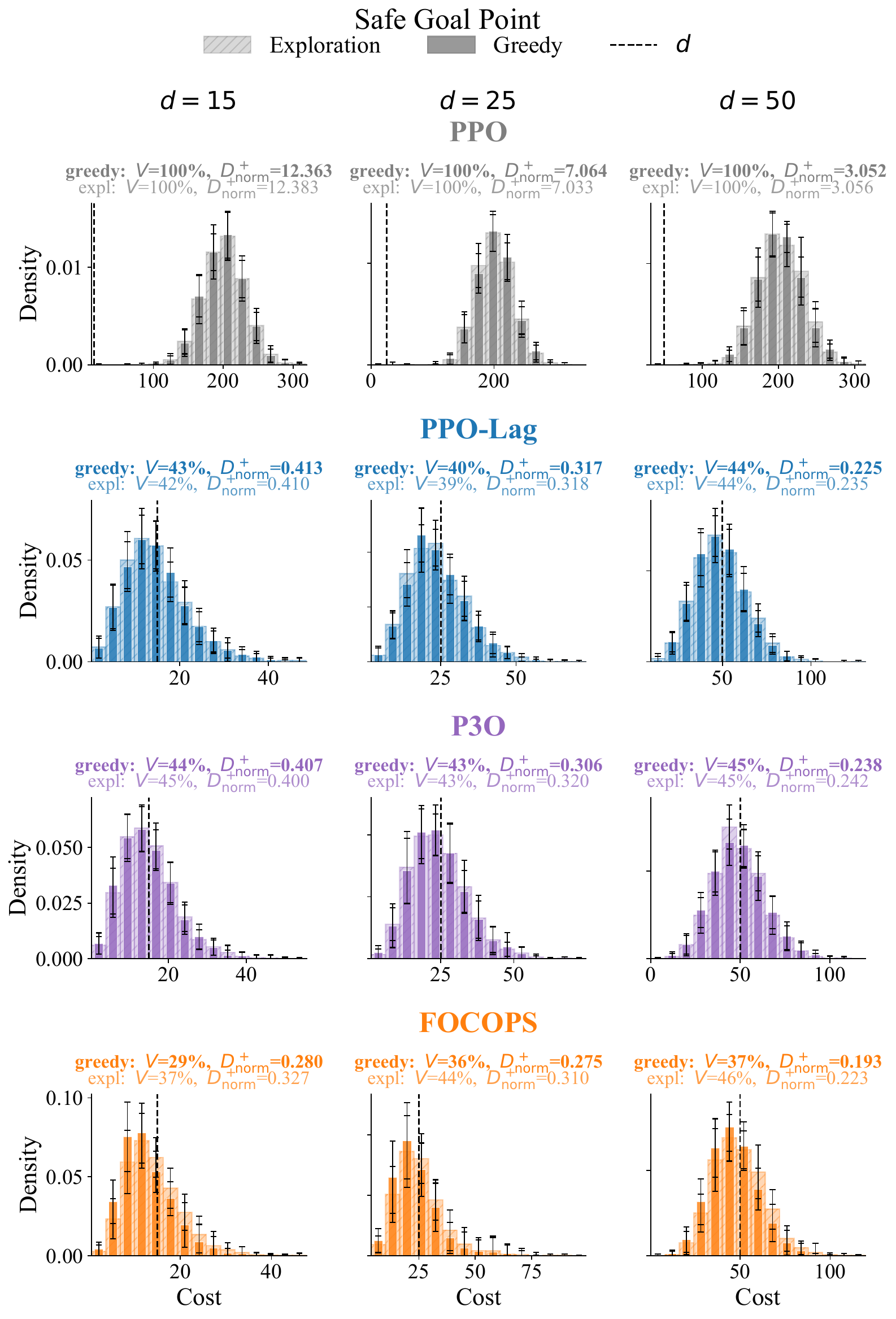}
    \caption{Episodic cost distributions for final policies with exploration noise on and off (greedy). Safe Goal Point (Level 1). Averaged over 30 seeds. Error bars denote $1\sigma$ over 30 seeds.}
    \label{fig:seed_hist_goal}
\end{figure}

\begin{figure}[H]
    \centering
    \includegraphics[width=1\linewidth]{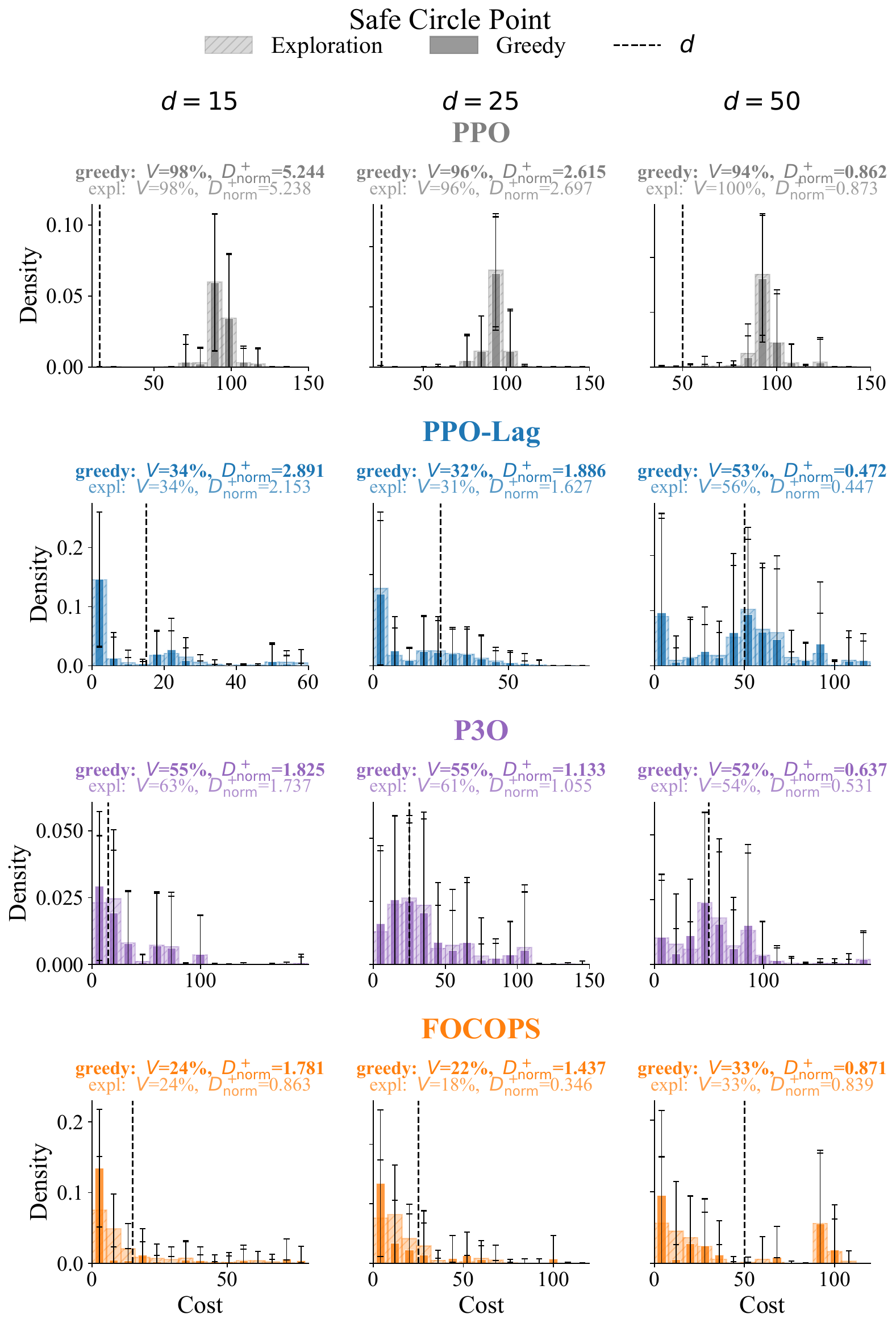}
    \caption{Episodic cost distributions for final policies with exploration noise on and off (greedy). Safe Circle Point (Level 1). Averaged over 30 seeds. Error bars denote $1\sigma$ over 30 seeds.}
    \label{fig:seed_hist_circle}
\end{figure}

\begin{figure}[H]
    \centering
    \includegraphics[width=1\linewidth]{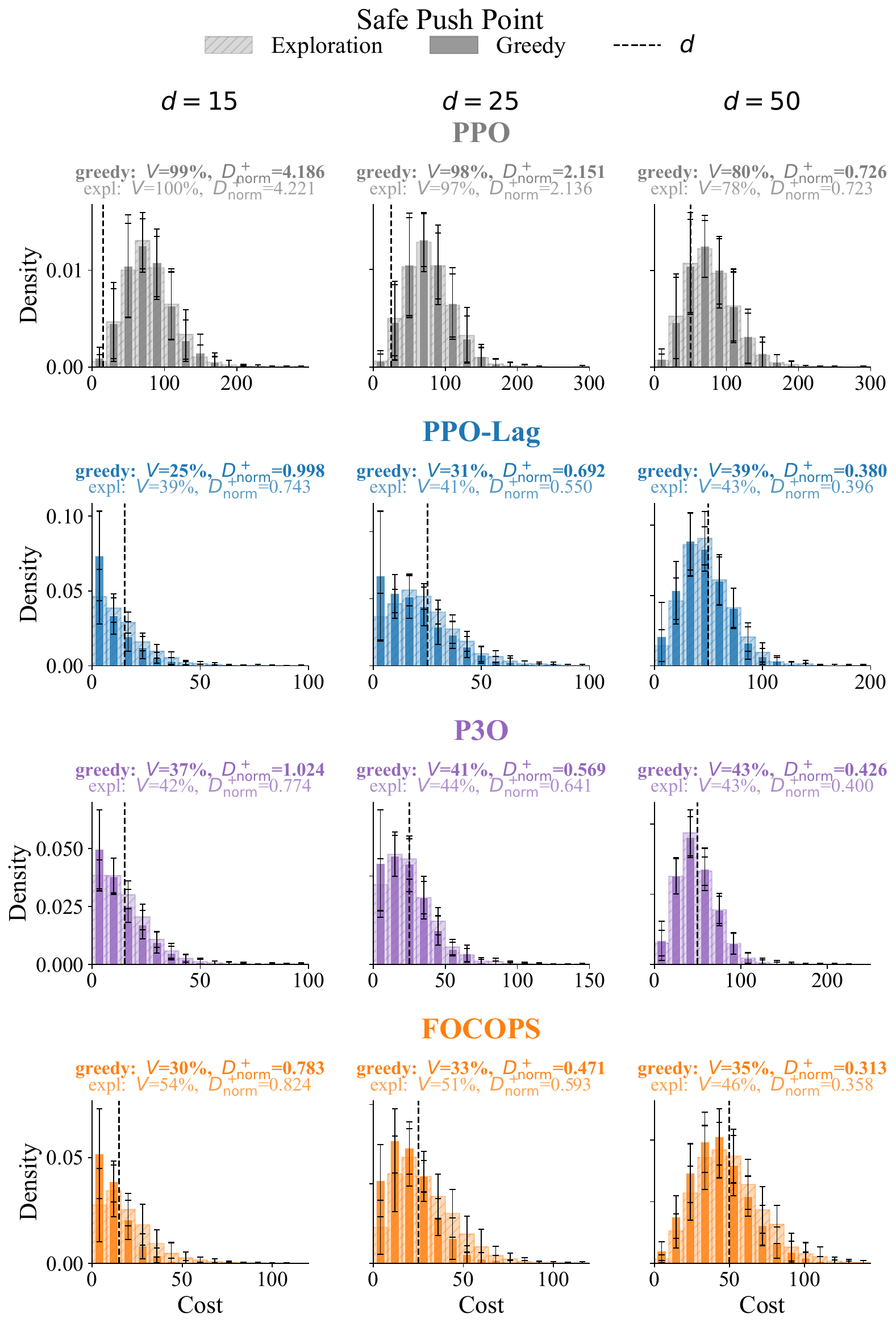}
    \caption{Episodic cost distributions for final policies with exploration noise on and off (greedy). Safe Push Point (Level 1). Averaged over 30 seeds. Error bars denote $1\sigma$ over 30 seeds.}
    \label{fig:seed_hist_push}
\end{figure}

\begin{figure}[H]
    \centering
    \includegraphics[width=1\linewidth]{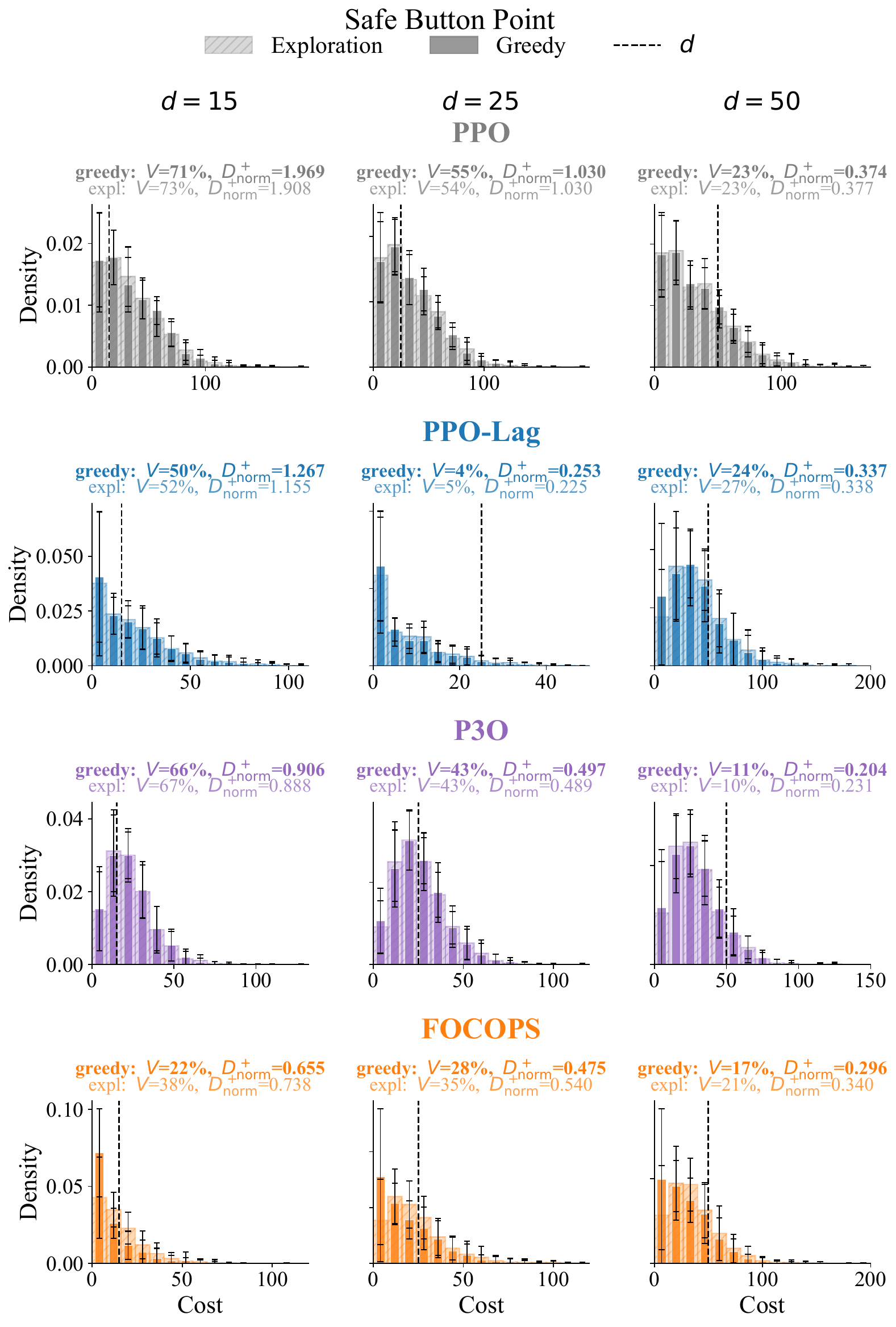}
    \caption{Episodic cost distributions for final policies with exploration noise on and off (greedy). Safe Button Point (Level 1). Averaged over 30 seeds. Error bars denote $1\sigma$ over 30 seeds.}
    \label{fig:seed_hist_button}
\end{figure}

All training curves plotted in Figures \ref{fig:seed_curves_goal}-\ref{fig:seed_curves_button} are smoothed using a moving average filter with a window of 5. The episodic distributions for final policies with exploration noise on or off (greedy) for each individual task and safety bound are shown in Figures \ref{fig:seed_hist_goal}-\ref{fig:seed_hist_button}, over $N=100$ rollouts. All metrics $X\in\{{\bar{R}, \bar{C}, D_{\text{norm}}, V, D^+_{\text{norm}}\}}$ are reported in Tables \ref{tab:per_bound_results_goal}-\ref{tab:per_bound_results_button} as $\frac{1}{K}\sum_{k=1}^{K} X_{(m,d,a,k)} \pm \sigma_k$, where $\sigma_k$ is the sample standard deviation for seed $k$, for $K=30$ seeds, for both over the entire course of training, and for the final greedy policy.

\begin{figure}[H]
    \centering
    \includegraphics[width=0.35\linewidth]{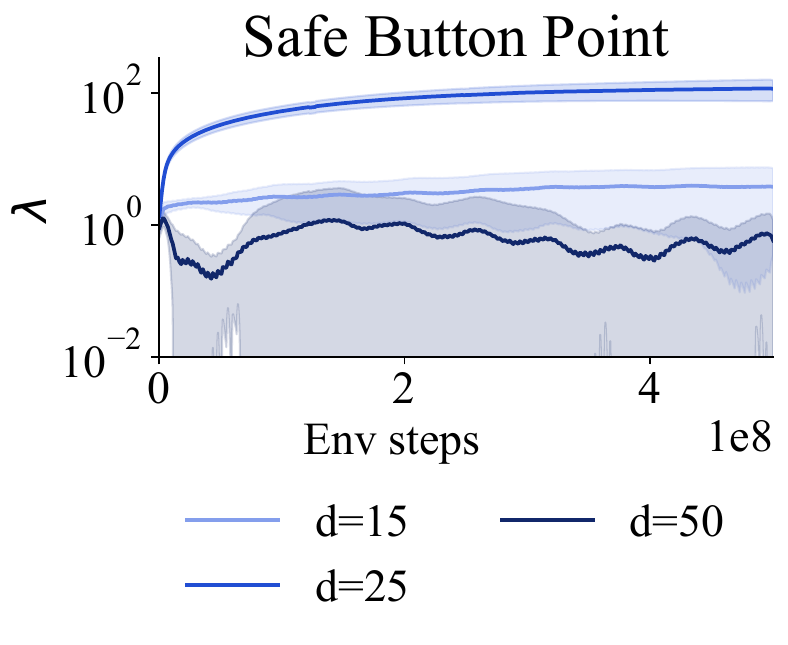}
    \caption{Lagrange multiplier $\lambda$ for PPO-Lag on Safe Button Point (Level 1), for safety bounds $d\in\{15,25,50\}$, averaged across 30 seeds. The shaded regions denote the $1\sigma$ over all seeds. The curve for $d=25$ shows that $\lambda$ grows very large compared to $d=15$ and $d=50$, suggesting that the policy for $d=25$ is frozen throughout the course of training.}
    \label{fig:button lambda}
\end{figure}

\paragraph{On the results for Safe Button Point}
The results for PPO-Lag on Safe Button Point in Table~\ref{tab:per_bound_results_button} show that, for $d=25$, the average reward $\bar{R}$ is negative at both training and for the final policy, and drops substantially further at the final policy performance ($-1.15\pm52.73$) compared to training ($-0.17\pm9.63$). We suspect this artifact is caused by a collapse in the cost, in which the Lagrange multiplier $\lambda$ grew too large during training, effectively freezing the policy into doing nothing as shown in Figure \ref{fig:seed_curves_button_25}. We confirm this by inspecting the corresponding Lagrange multiplier trajectory, shown in Figure~\ref{fig:button lambda}.

Furthermore, comparing the training and final policy columns for $\bar{R}$ and $D_{\text{norm}}$ across all algorithms on Safe Button Point in Table~\ref{tab:per_bound_results_button}, both metrics show a large drop from training to final policy performance. Rewards in this task come from actively pressing goal buttons \citep{tomilin2026crax}, which requires exploration or directed behavior. The final (greedy) policies of all evaluated safe RL algorithms appear to converge to a passive strategy resulting in low reward and low cost, possibly navigating the environment without pressing any buttons at all. This is consistent with the drop in $V$ observed at the final policy performance across all algorithms: pressing fewer buttons means fewer opportunities to trigger a hazard and violate the safety constraint.

\begin{table}[H]
\centering
\caption{Safety and performance evaluation per metric for Safe Goal Point (Level 1), for each individual $d\in\{15,25,50\}$. The numbers indicate mean $\pm 1\sigma$ over 30 seeds. \textbf{Bold text} indicates best result in that specific column, excluding PPO, as this is not a safe RL algorithm.}
\resizebox{\textwidth}{!}{%
\begin{tabular}{lccccccccccccc}
\toprule
\multirow{2}{*}{\textbf{Algorithm}} & \multicolumn{2}{c}{$\bar{R}$} & \multicolumn{2}{c}{$\bar{C}$} & \multicolumn{2}{c}{$D_{\text{norm}}$} & \multicolumn{2}{c}{$V$} & \multicolumn{2}{c}{$D_{\text{norm}^+}$} & \multicolumn{2}{c}{\textbf{Tier}} \\
\cmidrule(lr){2-3} \cmidrule(lr){4-5} \cmidrule(lr){6-7} \cmidrule(lr){8-9} \cmidrule(lr){10-11} \cmidrule(lr){12-13}
 & Train & Final (greedy) & Train & Final (greedy) & Train & Final (greedy) & Train & Final (greedy) & Train & Final (greedy) & Train & Final (greedy) \\
\midrule
\multicolumn{13}{l}{\textit{$d = 15$}} \\
\midrule
\textcolor{gray}{PPO}  & \textcolor{gray}{$72.45 \pm 1.50$}  & \textcolor{gray}{$72.05 \pm 1.59$} & \textcolor{gray}{$199.05 \pm 3.47$} & \textcolor{gray}{$200.39 \pm 5.23$} & \textcolor{gray}{$12.27 \pm 0.23$} & \textcolor{gray}{$12.36 \pm 0.35$} & \textcolor{gray}{$1.00 \pm 0.00$} & \textcolor{gray}{$1.00 \pm 0.00$} & \textcolor{gray}{$12.27 \pm 0.23$} & \textcolor{gray}{$12.36 \pm 0.35$} & \textcolor{gray}{0} & \textcolor{gray}{0} \\
PPO-Lag  & $62.46 \pm 2.34$ & $63.82 \pm 1.79$ & $14.84 \pm 0.25$ & $14.78 \pm 1.32$ & $\bm{-0.01 \pm 0.02}$ & $-0.01 \pm 0.09$ & $0.44 \pm 0.03$ & $0.43 \pm 0.08$ & $\bm{0.22 \pm 0.07}$ & $0.41 \pm 0.07$ & 2 & 2 \\
P3O      & $58.12 \pm 4.66$ & $63.15 \pm 1.96$ & $26.68 \pm 1.76$ & $14.80 \pm 1.21$ & $0.78 \pm 0.12$ & $-0.01 \pm 0.08$ & $0.62 \pm 0.03$ & $0.44 \pm 0.07$ & $1.29 \pm 0.16$ & $0.41 \pm 0.07$ & 0 & 2 \\
FOCOPS   & $\bm{68.86 \pm 0.41}$ & $\bm{73.73 \pm 0.67}$ & $\bm{15.79 \pm 0.05}$ & $\bm{12.65 \pm 2.19}$ & $0.05 \pm 0.00$ & $\bm{-0.16 \pm 0.15}$ & $\bm{0.32 \pm 0.03}$ & $\bm{0.29 \pm 0.14}$ & $0.44 \pm 0.07$ & $\bm{0.28 \pm 0.08}$ & 0 & 2 \\
\midrule
\multicolumn{13}{l}{\textit{$d = 25$}} \\
\midrule
\textcolor{gray}{PPO}  & \textcolor{gray}{$72.92 \pm 1.21$} & \textcolor{gray}{$72.35 \pm 1.33$} & \textcolor{gray}{$199.35 \pm 3.56$} & \textcolor{gray}{$201.53 \pm 5.87$} & \textcolor{gray}{$6.97 \pm 0.14$} & \textcolor{gray}{$7.06 \pm 0.23$} & \textcolor{gray}{$1.00 \pm 0.00$} & \textcolor{gray}{$1.00 \pm 0.00$} & \textcolor{gray}{$6.97 \pm 0.14$} & \textcolor{gray}{$7.06 \pm 0.24$} & \textcolor{gray}{0} & \textcolor{gray}{0} \\
PPO-Lag  & $64.70 \pm 1.65$ & $65.40 \pm 1.40$ & $24.79 \pm 0.36$ & $23.76 \pm 1.27$ & $\bm{-0.01 \pm 0.01}$ & $-0.05 \pm 0.05$ & $0.46 \pm 0.03$ & $0.40 \pm 0.06$ & $\bm{0.12 \pm 0.02}$ & $0.32 \pm 0.04$ & 2 & 2 \\
P3O      & $60.79 \pm 3.14$ & $64.00 \pm 1.91$ & $34.75 \pm 1.08$ & $24.01 \pm 2.40$ & $0.39 \pm 0.04$ & $-0.04 \pm 0.10$ & $0.59 \pm 0.02$ & $0.43 \pm 0.11$ & $0.70 \pm 0.08$ & $0.31 \pm 0.05$ & 0 & 2 \\
FOCOPS   & $\bm{70.44 \pm 0.35}$ & $\bm{74.17 \pm 2.12}$ & $\bm{25.49 \pm 0.04}$ & $\bm{23.53 \pm 6.57}$ & $0.02 \pm 0.00$ & $\bm{-0.06 \pm 0.26}$ & $\bm{0.37 \pm 0.03}$ & $\bm{0.36 \pm 0.21}$ & $0.22 \pm 0.02$ & $\bm{0.28 \pm 0.18}$ & 0 & 2 \\
\midrule
\multicolumn{13}{l}{\textit{$d = 50$}} \\
\midrule
\textcolor{gray}{PPO}  & \textcolor{gray}{$72.66 \pm 1.48$} & \textcolor{gray}{$72.34 \pm 1.34$} & \textcolor{gray}{$200.85 \pm 3.50$} & \textcolor{gray}{$202.53 \pm 5.50$} & \textcolor{gray}{$3.02 \pm 0.07$} & \textcolor{gray}{$3.05 \pm 0.11$} & \textcolor{gray}{$1.00 \pm 0.00$} & \textcolor{gray}{$1.00 \pm 0.00$} & \textcolor{gray}{$3.02 \pm 0.07$} & \textcolor{gray}{$3.05 \pm 0.11$} & \textcolor{gray}{0} & \textcolor{gray}{0} \\
PPO-Lag  & $67.32 \pm 1.46$ & $67.29 \pm 1.61$ & $49.87 \pm 0.22$ & $48.60 \pm 1.80$ & $\bm{0.00 \pm 0.00}$ & $-0.03 \pm 0.04$ & $0.49 \pm 0.04$ & $0.44 \pm 0.06$ & $\bm{0.06 \pm 0.01}$ & $0.23 \pm 0.03$ & 2 & 2 \\
P3O      & $64.86 \pm 2.71$ & $64.89 \pm 4.10$ & $57.01 \pm 0.92$ & $48.99 \pm 2.30$ & $0.14 \pm 0.02$ & $-0.02 \pm 0.05$ & $0.56 \pm 0.03$ & $0.45 \pm 0.07$ & $0.28 \pm 0.03$ & $0.24 \pm 0.04$ & 0 & 2 \\
FOCOPS   & $\bm{72.39 \pm 0.27}$ & $\bm{75.65 \pm 0.60}$ & $\bm{50.27 \pm 0.02}$ & $\bm{46.39 \pm 3.35}$ & $0.01 \pm 0.00$ & $\bm{-0.07 \pm 0.07}$ & $\bm{0.43 \pm 0.03}$ & $\bm{0.37 \pm 0.10}$ & $0.08 \pm 0.01$ & $\bm{0.19 \pm 0.04}$ &0 &2 \\
\bottomrule
\end{tabular}}
\label{tab:per_bound_results_goal}
\end{table}

\begin{table}[H]
\centering
\caption{Safety and performance evaluation per metric for Safe Circle Point (Level 1), for each individual $d\in\{15,25,50\}$. The numbers indicate mean $\pm 1\sigma$ over 30 seeds. \textbf{Bold text} indicates best result in that specific column, excluding PPO, as this is not a safe RL algorithm.}
\resizebox{\textwidth}{!}{%
\begin{tabular}{lccccccccccccc}
\toprule
\multirow{2}{*}{\textbf{Algorithm}} & \multicolumn{2}{c}{$\bar{R}$} & \multicolumn{2}{c}{$\bar{C}$} & \multicolumn{2}{c}{$D_{\text{norm}}$} & \multicolumn{2}{c}{$V$} & \multicolumn{2}{c}{$D_{\text{norm}^+}$} & \multicolumn{2}{c}{\textbf{Tier}} \\
\cmidrule(lr){2-3} \cmidrule(lr){4-5} \cmidrule(lr){6-7} \cmidrule(lr){8-9} \cmidrule(lr){10-11} \cmidrule(lr){12-13}
 & Train & Final (greedy) & Train & Final (greedy) & Train & Final (greedy) & Train & Final (greedy) & Train & Final (greedy) & Train & Final (greedy) \\
\midrule
\multicolumn{13}{l}{\textit{$d = 15$}} \\
\midrule
\textcolor{gray}{PPO}  & \textcolor{gray}{$234.91 \pm 17.41$} & \textcolor{gray}{$240.14 \pm 37.48$} & \textcolor{gray}{$93.02 \pm 5.40$} & \textcolor{gray}{$91.52 \pm 12.31$} & \textcolor{gray}{$5.20 \pm 0.36$} & \textcolor{gray}{$5.10 \pm 0.82$} & \textcolor{gray}{$0.99 \pm 0.01$} & \textcolor{gray}{$0.98 \pm 0.11$} & \textcolor{gray}{$5.24 \pm 0.35$} & \textcolor{gray}{$5.24 \pm 0.45$} & \textcolor{gray}{0} & \textcolor{gray}{0} \\
PPO-Lag   & $111.81 \pm 52.03$ & $100.74 \pm 80.28$ & $\bm{17.55 \pm 8.36}$ & $\bm{12.37 \pm 15.73}$ & $\bm{0.17 \pm 0.56}$ & $-0.18 \pm 1.05$ & $\bm{0.35 \pm 0.12}$ & $0.34 \pm 0.43$ & $2.02 \pm 1.10$ & $2.89 \pm 3.66$ & 0 & 2 \\
P3O       & $161.95 \pm 36.26$ & $151.60 \pm 69.61$ & $41.46 \pm 12.14$ & $29.86 \pm 25.48$ & $1.76 \pm 0.81$ & $0.99 \pm 1.70$ & $0.76 \pm 0.09$ & $0.55 \pm 0.38$ & $2.38 \pm 0.83$ & $1.83 \pm 1.83$ & 0 & 0 \\
FOCOPS    & $\bm{162.69 \pm 9.35}$  & $\bm{179.63 \pm 45.95}$ & $24.42 \pm 1.25$ & $11.55 \pm 21.07$ & $0.63 \pm 0.08$ & $\bm{-0.23 \pm 1.40}$ & $0.48 \pm 0.11$ & $\bm{0.24 \pm 0.42}$ & $\bm{1.65 \pm 0.50}$ & $\bm{1.78 \pm 1.54}$ & 0 & 2 \\
\midrule
\multicolumn{13}{l}{\textit{$d = 25$}} \\
\midrule
\textcolor{gray}{PPO}  & \textcolor{gray}{$230.53 \pm 20.00$} & \textcolor{gray}{$244.86 \pm 22.52$} & \textcolor{gray}{$92.04 \pm 7.72$} & \textcolor{gray}{$88.55 \pm 16.18$} & \textcolor{gray}{$2.68 \pm 0.31$} & \textcolor{gray}{$2.54 \pm 0.65$} & \textcolor{gray}{$0.97 \pm 0.05$} & \textcolor{gray}{$0.96 \pm 0.17$} & \textcolor{gray}{$2.78 \pm 0.23$} & \textcolor{gray}{$2.62 \pm 0.49$} & \textcolor{gray}{0} & \textcolor{gray}{0} \\
PPO-Lag   & $117.27 \pm 49.88$ & $128.72 \pm 75.68$ & $\bm{28.77 \pm 15.65}$ & $\bm{23.61 \pm 34.47}$ & $\bm{0.15 \pm 0.63}$ & $-0.06 \pm 1.38$ & $0.39 \pm 0.11$ & $0.31 \pm 0.40$ & $1.35 \pm 0.69$ & $1.89 \pm 2.14$ & 0 & 2 \\
P3O       & $152.18 \pm 40.39$ & $157.25 \pm 53.58$ & $47.50 \pm 9.04$ & $35.41 \pm 23.06$ & $0.90 \pm 0.36$ & $0.42 \pm 0.92$ & $0.74 \pm 0.09$ & $0.55 \pm 0.40$ & $1.32 \pm 0.34$ & $\bm{1.13 \pm 1.25}$ & 0 & 0 \\
FOCOPS    & $\bm{174.84 \pm 10.24}$ & $\bm{198.84 \pm 35.64}$ & $31.87 \pm 1.37$ & $19.19 \pm 30.20$ & $0.27 \pm 0.05$ & $\bm{-0.23 \pm 1.21}$ & $\bm{0.37 \pm 0.07}$ & $\bm{0.22 \pm 0.40}$ & $\bm{1.22 \pm 0.38}$ & $1.44 \pm 1.45$ & 0 & 2 \\
\midrule
\multicolumn{13}{l}{\textit{$d = 50$}} \\
\midrule
\textcolor{gray}{PPO}  & \textcolor{gray}{$234.20 \pm 16.73$} & \textcolor{gray}{$228.97 \pm 55.88$} & \textcolor{gray}{$91.60 \pm 8.90$} & \textcolor{gray}{$88.75 \pm 17.95$} & \textcolor{gray}{$0.83 \pm 0.18$} & \textcolor{gray}{$0.78 \pm 0.36$} & \textcolor{gray}{$0.96 \pm 0.08$} & \textcolor{gray}{$0.94 \pm 0.16$} & \textcolor{gray}{$0.89 \pm 0.10$} & \textcolor{gray}{$0.86 \pm 0.17$} & \textcolor{gray}{0} & \textcolor{gray}{0} \\
PPO-Lag   & $159.31 \pm 35.70$ & $172.76 \pm 69.93$ & $48.52 \pm 5.20$ & $48.39 \pm 27.42$ & $\bm{-0.03 \pm 0.10}$ & $-0.03 \pm 0.55$ & $0.48 \pm 0.05$ & $0.53 \pm 0.46$ & $\bm{0.53 \pm 0.33}$ & $\bm{0.47 \pm 0.54}$ & 2 & 1 \\
P3O       & $190.05 \pm 29.27$ & $166.43 \pm 70.37$ & $66.74 \pm 10.65$ & $56.41 \pm 31.54$ & $0.33 \pm 0.21$ & $0.13 \pm 0.63$ & $0.74 \pm 0.12$ & $0.52 \pm 0.43$ & $0.55 \pm 0.15$ & $0.64 \pm 0.59$ & 0 & 0 \\
FOCOPS    & $\bm{196.17 \pm 6.95}$  & $\bm{198.88 \pm 64.22}$ & $\bm{57.18 \pm 1.81}$ & $\bm{38.57 \pm 39.78}$ & $0.14 \pm 0.04$ & $\bm{-0.23 \pm 0.80}$ & $\bm{0.47 \pm 0.04}$ & $\bm{0.33 \pm 0.47}$ & $0.80 \pm 0.08$ & $0.87 \pm 0.41$ & 0 & 2 \\
\bottomrule
\end{tabular}}
\label{tab:per_bound_results_circle}
\end{table}

\begin{table}[H]
\centering
\caption{Safety and performance evaluation per metric for Safe Push Point (Level 1), for each individual $d\in\{15,25,50\}$. The numbers indicate mean $\pm 1\sigma$ over 30 seeds. \textbf{Bold text} indicates best result in that specific column, excluding PPO, as this is not a safe RL algorithm.}
\resizebox{\textwidth}{!}{%
\begin{tabular}{lccccccccccccc}
\toprule
\multirow{2}{*}{\textbf{Algorithm}} & \multicolumn{2}{c}{$\bar{R}$} & \multicolumn{2}{c}{$\bar{C}$} & \multicolumn{2}{c}{$D_{\text{norm}}$} & \multicolumn{2}{c}{$V$} & \multicolumn{2}{c}{$D_{\text{norm}^+}$} & \multicolumn{2}{c}{\textbf{Tier}} \\
\cmidrule(lr){2-3} \cmidrule(lr){4-5} \cmidrule(lr){6-7} \cmidrule(lr){8-9} \cmidrule(lr){10-11} \cmidrule(lr){12-13}
 & Train & Final (greedy) & Train & Final (greedy) & Train & Final (greedy) & Train & Final (greedy) & Train & Final (greedy) & Train & Final (greedy) \\
\midrule
\multicolumn{13}{l}{\textit{$d = 15$}} \\
\midrule
\textcolor{gray}{PPO}  & \textcolor{gray}{$57.07 \pm 9.99$}  & \textcolor{gray}{$64.79 \pm 10.41$} & \textcolor{gray}{$85.43 \pm 15.02$} & \textcolor{gray}{$77.26 \pm 15.05$} & \textcolor{gray}{$4.70 \pm 1.00$} & \textcolor{gray}{$4.15 \pm 1.00$} & \textcolor{gray}{$1.00 \pm 0.00$} & \textcolor{gray}{$0.99 \pm 0.01$} & \textcolor{gray}{$4.70 \pm 1.00$} & \textcolor{gray}{$4.19 \pm 1.00$} & \textcolor{gray}{0} & \textcolor{gray}{0} \\
PPO-Lag  & $27.13 \pm 15.82$ & $33.84 \pm 22.98$ & $\bm{15.23 \pm 0.15}$ & $\bm{10.69 \pm 3.45}$ & $\bm{0.02 \pm 0.01}$ & $\bm{-0.29 \pm 0.23}$ & $0.44 \pm 0.04$ & $\bm{0.25 \pm 0.12}$ & $\bm{0.15 \pm 0.07}$ & $1.00 \pm 0.83$ & 0 & 2 \\
P3O      & $38.51 \pm 17.15$ & $52.49 \pm 18.49$ & $20.44 \pm 1.16$ & $14.93 \pm 3.44$ & $0.36 \pm 0.08$ & $-0.00 \pm 0.23$ & $0.71 \pm 0.08$ & $0.37 \pm 0.09$ & $0.54 \pm 0.11$ & $1.02 \pm 0.64$ & 0 & 2 \\
FOCOPS   & $\bm{56.53 \pm 11.59}$ & $\bm{63.78 \pm 22.21}$ & $16.18 \pm 0.27$ & $12.52 \pm 4.87$ & $0.08 \pm 0.02$ & $-0.17 \pm 0.32$ & $\bm{0.42 \pm 0.05}$ & $0.30 \pm 0.15$ & $0.37 \pm 0.04$ & $\bm{0.78 \pm 0.88}$ & 0 & 2 \\
\midrule
\multicolumn{13}{l}{\textit{$d = 25$}} \\
\midrule
\textcolor{gray}{PPO}  & \textcolor{gray}{$57.13 \pm 9.60$} & \textcolor{gray}{$64.15 \pm 10.40$} & \textcolor{gray}{$85.11 \pm 14.41$} & \textcolor{gray}{$77.52 \pm 14.28$} & \textcolor{gray}{$2.40 \pm 0.58$} & \textcolor{gray}{$2.10 \pm 0.57$} & \textcolor{gray}{$1.00 \pm 0.00$} & \textcolor{gray}{$0.98 \pm 0.03$} & \textcolor{gray}{$2.40 \pm 0.58$} & \textcolor{gray}{$2.15 \pm 0.56$} & \textcolor{gray}{0} & \textcolor{gray}{0} \\
PPO-Lag  & $33.45 \pm 17.36$ & $42.98 \pm 22.80$ & $\bm{25.18 \pm 0.29}$ & $\bm{20.49 \pm 4.85}$ & $\bm{0.01 \pm 0.01}$ & $\bm{-0.18 \pm 0.19}$ & $0.45 \pm 0.04$ & $\bm{0.31 \pm 0.11}$ & $\bm{0.12 \pm 0.10}$ & $0.69 \pm 0.43$ &0 & 2\\
P3O      & $39.50 \pm 17.30$ & $50.25 \pm 18.56$ & $29.66 \pm 2.10$ & $23.55 \pm 4.51$ & $0.19 \pm 0.08$ & $-0.06 \pm 0.18$ & $0.64 \pm 0.12$ & $0.41 \pm 0.10$ & $0.34 \pm 0.15$ & $0.57 \pm 0.12$ &0 & 2 \\
FOCOPS   & $\bm{63.27 \pm 12.00}$ & $\bm{76.30 \pm 12.88}$ & $25.49 \pm 0.34$ & $21.43 \pm 6.06$ & $0.02 \pm 0.01$ & $-0.14 \pm 0.24$ & $\bm{0.39 \pm 0.03}$ & $0.33 \pm 0.13$ & $0.23 \pm 0.04$ & $\bm{0.47 \pm 0.42}$ & 0& 2\\
\midrule
\multicolumn{13}{l}{\textit{$d = 50$}} \\
\midrule
\textcolor{gray}{PPO}    & \textcolor{gray}{$55.57 \pm 10.01$} & \textcolor{gray}{$63.36 \pm 10.15$} & \textcolor{gray}{$86.52 \pm 16.04$} & \textcolor{gray}{$77.98 \pm 16.49$} & \textcolor{gray}{$0.73 \pm 0.32$} & \textcolor{gray}{$0.56 \pm 0.33$} & \textcolor{gray}{$0.99 \pm 0.05$} & \textcolor{gray}{$0.80 \pm 0.15$} & \textcolor{gray}{$0.73 \pm 0.32$} & \textcolor{gray}{$0.73 \pm 0.26$} & \textcolor{gray}{0} & \textcolor{gray}{0} \\
PPO-Lag  & $42.73 \pm 15.06$ & $51.73 \pm 17.17$ & $\bm{48.78 \pm 3.17}$ & $\bm{46.31 \pm 5.81}$ & $\bm{-0.02 \pm 0.06}$ & $-0.07 \pm 0.12$ & $0.39 \pm 0.17$ & $0.39 \pm 0.09$ & $0.10 \pm 0.08$ & $0.38 \pm 0.09$ & 2 & 2 \\
P3O      & $43.69 \pm 14.52$ & $52.02 \pm 15.01$ & $51.35 \pm 2.40$ & $49.26 \pm 3.03$ & $0.03 \pm 0.05$ & $-0.01 \pm 0.06$ & $0.48 \pm 0.16$ & $0.43 \pm 0.06$ & $0.17 \pm 0.14$ & $0.43 \pm 0.12$ & 0 & 2 \\
FOCOPS   & $\bm{71.08 \pm 8.84}$ & $\bm{81.92 \pm 7.79}$ & $48.83 \pm 1.08$ & $44.14 \pm 4.86$ & $-0.02 \pm 0.02$ & $\bm{-0.12 \pm 0.10}$ & $\bm{0.33 \pm 0.09}$ & $\bm{0.35 \pm 0.10}$ & $\bm{0.09 \pm 0.01}$ & $\bm{0.31 \pm 0.06}$ & 2 & 2 \\
\bottomrule
\end{tabular}}
\label{tab:per_bound_results_push}
\end{table}

\begin{table}[H]
\centering
\caption{Safety and performance evaluation per metric for Safe Button Point (Level 1), for each individual $d\in\{15,25,50\}$. The numbers indicate mean $\pm 1\sigma$ over 30 seeds. \textbf{Bold text} indicates best result in that specific column, excluding PPO, as this is not a safe RL algorithm.}
\resizebox{\textwidth}{!}{%
\begin{tabular}{lccccccccccccc}
\toprule
\multirow{2}{*}{\textbf{Algorithm}} & \multicolumn{2}{c}{$\bar{R}$} & \multicolumn{2}{c}{$\bar{C}$} & \multicolumn{2}{c}{$D_{\text{norm}}$} & \multicolumn{2}{c}{$V$} & \multicolumn{2}{c}{$D_{\text{norm}^+}$} & \multicolumn{2}{c}{\textbf{Tier}} \\
\cmidrule(lr){2-3} \cmidrule(lr){4-5} \cmidrule(lr){6-7} \cmidrule(lr){8-9} \cmidrule(lr){10-11} \cmidrule(lr){12-13}
 & Train & Final (greedy) & Train & Final (greedy) & Train & Final (greedy) & Train & Final (greedy) & Train & Final (greedy) & Train & Final (greedy) \\
\midrule
\multicolumn{13}{l}{\textit{$d = 15$}} \\
\midrule
\textcolor{gray}{PPO}  & \textcolor{gray}{$158.70 \pm 8.70$} & \textcolor{gray}{$128.43 \pm 18.09$} & \textcolor{gray}{$351.56 \pm 30.21$} & \textcolor{gray}{$34.31 \pm 5.14$} & \textcolor{gray}{$22.44 \pm 2.01$} & \textcolor{gray}{$1.29 \pm 0.34$} & \textcolor{gray}{$1.00 \pm 0.00$} & \textcolor{gray}{$0.71 \pm 0.10$} & \textcolor{gray}{$22.44 \pm 2.01$} & \textcolor{gray}{$1.97 \pm 0.33$} & \textcolor{gray}{0} & \textcolor{gray}{0} \\
PPO-Lag  & $121.04 \pm 13.19$ & $13.34 \pm 15.96$ & $\bm{16.02 \pm 1.27}$ & $21.22 \pm 10.16$ & $\bm{0.07 \pm 0.08}$ & $0.41 \pm 0.68$ & $\bm{0.47 \pm 0.11}$ & $0.50 \pm 0.23$ & $\bm{0.34 \pm 0.11}$ & $1.27 \pm 0.52$ & 0 & 0 \\
P3O      & $119.25 \pm 9.73$  & $\bm{50.96 \pm 22.92}$ & $44.49 \pm 5.53$ & $22.55 \pm 5.71$ & $1.97 \pm 0.37$ & $0.50 \pm 0.38$ & $1.00 \pm 0.00$ & $0.66 \pm 0.17$ & $1.97 \pm 0.37$ & $0.91 \pm 0.25$ & 0 & 0 \\
FOCOPS   & $\bm{126.48 \pm 7.90}$ & $2.96 \pm 5.69$ & $20.51 \pm 0.56$ & $\bm{10.11 \pm 7.67}$ & $0.37 \pm 0.04$ & $\bm{-0.33 \pm 0.51}$ & $0.80 \pm 0.06$ & $\bm{0.22 \pm 0.22}$ & $0.48 \pm 0.02$ & $\bm{0.66 \pm 0.36}$ & 0 & 2 \\
\midrule
\multicolumn{13}{l}{\textit{$d = 25$}} \\
\midrule
\textcolor{gray}{PPO}  & \textcolor{gray}{$159.05 \pm 6.10$} & \textcolor{gray}{$131.43 \pm 14.21$} & \textcolor{gray}{$349.05 \pm 36.43$} & \textcolor{gray}{$33.98 \pm 3.94$} & \textcolor{gray}{$12.96 \pm 1.46$} & \textcolor{gray}{$0.36 \pm 0.16$} & \textcolor{gray}{$1.00 \pm 0.00$} & \textcolor{gray}{$0.55 \pm 0.08$} & \textcolor{gray}{$12.96 \pm 1.46$} & \textcolor{gray}{$1.03 \pm 0.16$} & \textcolor{gray}{0} & \textcolor{gray}{0} \\
PPO-Lag  & $-0.17 \pm 9.63$   & $-1.15 \pm 52.73$ & $\bm{2.33 \pm 0.46}$ & $\bm{7.38 \pm 4.11}$ & $\bm{-0.91 \pm 0.02}$ & $\bm{-0.70 \pm 0.16}$ & $\bm{0.00 \pm 0.00}$ & $\bm{0.04 \pm 0.05}$ & $0.77 \pm 0.32$ & $\bm{0.25 \pm 0.10}$ & 2 & 2 \\
P3O      & $116.64 \pm 10.90$ & $\bm{63.73 \pm 23.86}$ & $54.62 \pm 5.23$ & $25.04 \pm 5.14$ & $1.18 \pm 0.21$ & $0.00 \pm 0.21$ & $0.99 \pm 0.01$ & $0.43 \pm 0.15$ & $1.20 \pm 0.21$ & $0.50 \pm 0.12$ & 0 & 2 \\
FOCOPS   & $\bm{149.51 \pm 7.69}$ & $8.14 \pm 11.04$ & $26.24 \pm 0.35$ & $18.63 \pm 9.98$ & $0.05 \pm 0.01$ & $-0.25 \pm 0.40$ & $0.41 \pm 0.05$ & $0.28 \pm 0.23$ & $\bm{0.28 \pm 0.05}$ & $0.48 \pm 0.20$ & 0 & 2 \\
\midrule
\multicolumn{13}{l}{\textit{$d = 50$}} \\
\midrule
\textcolor{gray}{PPO}  & \textcolor{gray}{$158.01 \pm 5.82$} & \textcolor{gray}{$130.56 \pm 13.11$} & \textcolor{gray}{$353.01 \pm 26.43$} & \textcolor{gray}{$33.55 \pm 5.15$} & \textcolor{gray}{$6.06 \pm 0.53$} & \textcolor{gray}{$-0.33 \pm 0.10$} & \textcolor{gray}{$1.00 \pm 0.00$} & \textcolor{gray}{$0.23 \pm 0.08$} & \textcolor{gray}{$6.06 \pm 0.53$} & \textcolor{gray}{$0.37 \pm 0.10$} & \textcolor{gray}{0} & \textcolor{gray}{2} \\
PPO-Lag  & $148.02 \pm 12.79$ & $44.48 \pm 31.20$ & $\bm{47.39 \pm 1.97}$ & $37.15 \pm 13.81$ & $\bm{-0.05 \pm 0.04}$ & $-0.26 \pm 0.28$ & $\bm{0.30 \pm 0.11}$ & $0.24 \pm 0.20$ & $0.14 \pm 0.08$ & $0.34 \pm 0.14$ & 2 & 2 \\
P3O      & $122.83 \pm 10.21$ & $\bm{80.36 \pm 23.98}$ & $75.26 \pm 4.41$ & $\bm{28.33 \pm 6.55}$ & $0.51 \pm 0.09$ & $\bm{-0.43 \pm 0.13}$ & $0.93 \pm 0.03$ & $\bm{0.11 \pm 0.08}$ & $0.56 \pm 0.09$ & $\bm{0.20 \pm 0.07}$ & 0 & 2 \\
FOCOPS   & $\bm{174.33 \pm 6.08}$ & $20.82 \pm 23.66$ & $50.02 \pm 0.12$ & $29.85 \pm 11.86$ & $0.00 \pm 0.00$ & $-0.40 \pm 0.24$ & $0.41 \pm 0.03$ & $0.17 \pm 0.15$ & $\bm{0.08 \pm 0.01}$ & $0.30 \pm 0.11$ & 2 & 2 \\
\bottomrule
\end{tabular}}
\label{tab:per_bound_results_button}
\end{table}

\end{document}